\documentclass[pdflatex,sn-mathphys-num]{sn-jnl}

\newcommand{\agenttool}[1]{\texttt{#1}}

\usepackage{rotating}
\usepackage{graphicx}%
\usepackage{multirow}%
\usepackage{amsmath,amssymb,amsfonts}%
\usepackage{mathrsfs}%
\usepackage[title]{appendix}%
\usepackage{xcolor}%
\usepackage{textcomp}%
\usepackage{manyfoot}%
\usepackage{booktabs}%
\usepackage{fancyhdr}
\usepackage{longtable}
\usepackage{algorithm}%
\usepackage{algorithmicx}%
\usepackage{algpseudocode}%
\usepackage{listings}%
\newcommand{\mycomment}[1]{}
\usepackage{tabularx}
\usepackage{booktabs}
\usepackage{array}
\usepackage{xcolor}
\usepackage{multirow}
\usepackage[most]{tcolorbox}
\usepackage{makecell}
\usepackage{booktabs, makecell}
\usepackage{pdfpages}
\usepackage{pgffor}
\usepackage{caption}
\usepackage{float}
\newcommand{\catsep}{\enspace$\cdot$\enspace}
\usepackage{booktabs}
\usepackage{xltabular}

\theoremstyle{thmstyleone}%
\theoremstyle{thmstyletwo}%

\theoremstyle{thmstylethree}%

\begin{document}

\title[Agentic AI for Healthcare Claim Denial Analysis: A Policy-Grounded Framework with Privacy Safeguard]{Agentic AI for Healthcare Claim Denial Analysis: A Policy-Grounded Framework for Privacy Preservation}


\author*[1]{\fnm{Sofia} \sur{Campos}} \email{sofia.campos@randylabs.com}

\author[1]{\fnm{Carlos} \sur{Martins}}

\author[1]{\fnm{Pedro} \sur{Pires}}

\author[2,3]{\fnm{João} \sur{Gama}}

\author[2,3]{\fnm{Bruno} \sur{Veloso}}

\author[1]{\fnm{Pedro} \sur{Santos}}

\affil[1]{\orgname{Randy Labs}, \country{Portugal}}

\affil*[2]{\orgname{University of Porto}, \country{Portugal}}

\affil[3]{\orgname{INESC TEC}, \country{Portugal}}

\mycomment{
\author*{\fnm{Sofia} \sur{Campos}}
\equalcont{These authors contributed equally to this work.}

\author{\fnm{Carlos} \sur{Martins}}
\equalcont{These authors contributed equally to this work.}

\author{\fnm{Pedro} \sur{Pires}}
\equalcont{These authors contributed equally to this work.}

\author{\fnm{João} \sur{Gama}}
\equalcont{These authors contributed equally to this work.}

\author{\fnm{Bruno} \sur{ Veloso}}
\equalcont{These authors contributed equally to this work.}

\author{\fnm{Pedro} \sur{Santos}}
\equalcont{These authors contributed equally to this work.}
}


\abstract{In the healthcare sector, claim denials impose significant financial and administrative burdens, emphasizing the need for automated and trustworthy decision-support systems. This paper proposes a modular Agentic Artificial Intelligence (AI) framework for healthcare claim denial analysis, integrating specialized agents for denial risk prediction, risk factor identification, diagnosis–procedure–medication code relationship analysis, and counterfactual analysis. The framework integrates predictive analytics, graph-based learning, a Hybrid Contextual Retrieval-Augmented Generation (RAG) pipeline grounded in healthcare policy documents, and a dedicated HIPAA Compliance Agent to support privacy protection. Evaluated on a synthetic healthcare claims dataset, the framework achieved 98\% traceability of reported numerical findings, validated 214 of 216 policy citations, and removed all 280 identified sensitive identifiers without modifying the underlying analysis. These findings pave the way for further research into Agentic AI as a framework for explainable, policy-aware, and privacy-preserving healthcare claim denial management.}




\maketitle

\section{Introduction}\label{sec1}

AI has evolved rapidly, driving widespread adoption across industries and transforming digital systems into core components of society  \cite{Kamath2024}.

\emph{Large Language Models} (LLMs) have emerged as powerful tools for text generation and complex problem-solving. However, these standalone models exhibit limitations such as hallucinations \cite{Pradeep2024}, restricted context length \cite{kenton2019}, limited collaboration \cite{lamalfa2025}, and lack of external integration capabilities \cite{li2024}, which constrain their adaptability to real-world workflows \cite{annepaka2025, Peykani2025}.

These constraints have fostered a new paradigm, the “Agentic Era", where autonomous, goal-driven AI agents collaborate dynamically and reason strategically.
In healthcare, Agentic AI represents a promising approach to overcoming challenges in efficiency, resource allocation, and administrative burden \cite{Nalan2025}.

One domain that exemplifies this necessity is claim denial management, which encompasses the process of preventing, analyzing, and resolving payment denials submitted by healthcare providers to insurers. Every year, healthcare providers lose significant revenue from preventable claim denials, as manual workflows and growing data volumes make denial management increasingly complex, leading to substantial financial losses and delays in care \cite{Hoagland2024}. Effective automation is therefore critical to improve operational efficiency and healthcare accessibility \cite{Li2023multimodal}.
However, deploying such systems introduces challenges in privacy, accuracy, and regulatory compliance \cite{reddy2024}. AI frameworks must adhere to strict legal standards such as the HIPAA  and the \emph{General Data Protection Regulation} (GDPR). Despite theoretical progress, empirical research on the integration of RAG within compliant healthcare systems remains limited \cite{Ajit2025}.

To bridge this gap, we propose a modular Agentic AI framework for healthcare claim denial management, integrating a Hybrid Contextual Retrieval-Augmented Generation (RAG) pipeline with a dedicated HIPAA Compliance Agent.

We introduce the first operational blueprint for Agentic AI in healthcare regulation, featuring a modular design adaptable to GDPR and national contexts, and unifying explainability, privacy, and compliance into a single framework, with the following main contributions:

\begin{enumerate}
  \renewcommand{\labelenumi}{(\arabic{enumi})}
  \setlength{\itemsep}{5pt}

  \item Implementation of a \textbf{multi-agent architecture} for healthcare claim denial analysis, orchestrated by a Coordinator Agent and comprising four specialist agents dedicated to denial prediction, risk-driver identification, code-combination analysis, and counterfactual causality.

  \item Integration of \textbf{Machine Learning} models, \textit{\textbf{Graph Neural Networks (GNNs)}}, and \textbf{Counterfactual Analysis} to support denial risk prediction, uncover relationships between diagnosis, procedure, and medication codes associated with claim denials, and evaluate hypothetical claim modifications that may reduce predicted denial risk.
  
  \item Incorporation of a \textbf{Hybrid Contextual RAG} pipeline to retrieve and contextualize relevant payer policies and supporting documentation, grounding AI-generated insights in domain-specific evidence.

  \item Consolidation via \textbf{HIPAA} Compliance Agent, supporting the exploration of potential alternative claim scenarios and the review of generated reports for exposed sensitive information and potential privacy violations.
\end{enumerate}

Finally, the paper is organized as follows: Section~\ref{sec:literature} clarifies theoretical concepts and reviews related work in the scope; Section~\ref{sec:methodology} outlines the proposed methodology and framework design; Section ~\ref{sec:data} describes the data and exploratory data analysis conducted; Section~\ref{sec:experiments} presents the experimental results and their implications; and Section~\ref{sec:conclusions} concludes with key findings, limitations, and directions for future research.

\section{Related Work} \label{sec:literature}
This section reviews related work supporting the proposed approach, covering challenges in claim denial management, the rise of Agentic AI, and the integration of RAG mechanisms. Finally, it concludes with the role of Agentic AI in healthcare, outlining current applications, limitations, and research opportunities.

\subsection{Claim Denial Management}
Claim Denial Management remains hindered by multiple inefficiencies, including:
\begin{itemize}
\setlength{\itemsep}{5pt}
\item \textbf{Frequent reimbursement denials} that strain finances and operations, often caused by coding errors, inadequate coverage, or administrative issues.
\item \textbf{Rising medical data volumes} and persistent manual review processes, still used by 48\% of providers \cite{experian2025}.
\item \textbf{Evolving payer policies} and pre-authorization requirements that create additional reimbursement barriers.
\item \textbf{Process and system misalignments} generating inefficiencies, revenue risks, and higher costs.
\item \textbf{Restrictive regulatory frameworks} that protect health data privacy but complicate interoperability and standardization.
\end{itemize}

These challenges require robust and intelligent frameworks that combine automation, regulatory compliance, fraud detection, and patient-centric design.

Consequently, healthcare providers seek to automate claim workflows and stakeholder interactions while maintaining strict compliance with data protection regulations, an enduring challenge in the healthcare sector. Overcoming these complexities requires innovative solutions, as regulatory frameworks vary widely across regions, reflecting different legal and policy contexts, as detailed by \cite{AI_Act_2024, AIRegulation, Mennella2024}.

\subsection{Conceptual AI Framework}

Agentic AI represents a paradigm shift beyond conventional AI, evolving from isolated agents into collaborative, multi-agent ecosystems capable of decomposing and executing complex goals \cite{Acharya2025}. These systems comprise orchestrated agents that perceive their environment and interact through tools, APIs, or shared workspaces.
An AI framework defines how agents operate, set goals, make decisions, and interact intelligently with tools and environments. Thus, its cognitive architecture typically includes three key components: the \textbf{Model}, the \textbf{Orchestration Layer}, and the \textbf{Tools}.
\begin{enumerate}
\item \textbf{Model}: any \emph{Language Model} (e.g., LLM) capable of instruction-based reasoning and logical inference.
\item \textbf{Orchestration Layer}: manages the agent’s information intake, reasoning, and decision cycle \cite{Wiesinger2024}.

Core functions include:
\begin{itemize}
\item \emph{Planning and Decomposition} — breaking down complex tasks into sub-goals via mechanisms such as \emph{Chain-of-Thought} (CoT) and \emph{Tree-of-Thought} (ToT).
\item \emph{Reflection and Refinement} — enabling self-criticism and performance improvement, as in ReAct, Reinforcement Learning, among others.
\item \emph{Memory} — mechanisms for storing and retrieving information, spanning \emph{Short-Term} and \emph{Long-Term Memory}.
\end{itemize}
\item \textbf{Tools}: keys that enable LLM-based agents to “think outside the world'', whose integration significantly enhances the capabilities of agentic models. A prominent example of such integration is RAG, a technique that equips LLMs with access to external knowledge bases at inference time, which enriches LLM outputs with up-to-date and domain-relevant information from structured (e.g., SQL databases) and unstructured (e.g., text, PDFs) sources \cite{lin2024}.
RAG mitigates hallucinations, static knowledge, and limited domain generalization, grounding responses in retrieved top-$k$ documents \cite{wang2022gpl}.
\end{enumerate}

\subsection{Agentic AI in Healthcare}
Recent advances in Agentic AI have sparked considerable interest in healthcare for their ability to autonomously execute complex, multi-step clinical workflows. Applications have emerged across several tasks, including medical report generation \cite{sudarshan2024}, clinical service delivery \cite{chen2025clinical}, and interpretation of cognitive patterns in clinical notes \cite{tian2025}.

Within this scope, Tzanis and Klontzas \cite{tzanis2025} introduced \emph{mAIstro}, an open-source multi-agent system for automating AI model development in medical imaging and tabular data, which demonstrated a significant reduction in development time and technical barriers. However, it lacks RAG integration and HIPAA-compliant mechanisms for data privacy and regulatory adherence.
Similarly, Neupane et al. \cite{neupane2025} proposed a HIPAA-compliant Agentic AI framework using \emph{Attribute-Based Access Control} (ABAC), automated \emph{Protected Health Information} (PHI) removal, and audit logging to ensure secure data handling. While enabling privacy-preserving decision-making, the system also omits RAG capabilities, limiting dynamic knowledge retrieval.

Recent studies demonstrate that RAG frameworks effectively retrieve relevant documents and generate insightful responses, proving highly suitable for medical research applications. However, despite a solid theoretical foundation, empirical evidence on their use in healthcare remains limited \cite{Ajit2025}. This limitation motivates our investigation into the extent to which RAG can improve explainability and regulatory compliance in healthcare by grounding model outputs in source documents.

\section{Methods} \label{sec:methodology}

The present section outlines the methodology implemented in this study, structured around the following Research Question (RQ), which guides the overall design and ensures alignment with the research objectives:

\begin{enumerate}
\item[\textbf{RQ}] \label{rq} To what extent can an Agentic AI framework integrating specialist agents, Hybrid Contextual RAG, and a dedicated compliance layer enhance the faithfulness, contextual grounding, transparency, and privacy protection of AI-generated insights?
\end{enumerate}

\subsection{Model Selection \& Architectural Design}

To test our framework, we selected the \textit{gpt-oss-120b} LLM model for its balance between reasoning quality and efficiency, native function calling, and structured outputs. Served through Groq’s low-latency, OpenAI-compatible API, it powers the coordinator and four specialist agents, including a HIPAA compliance reviewer. All agents use the same client with a temperature of 0.1 for more consistent outputs. A complete claim analysis takes approximately \textbf{18 seconds} on average.

The workflow implements three retry layers: transport-level (up to 8 attempts), agent-level (2–3 retries), and run-level (up to 3 retries). LLM requests have a 600-second timeout and a shared budget of 50 requests per run.

Ergo, our study evaluates whether augmenting agentic reasoning with healthcare-specific documentation, such as denial trends, official payer policies, for instance, coverage guidelines, eligibility criteria, and justification codes, can enhance the explainability and trustworthiness of AI-generated insights, providing clearer, context-aware outputs for healthcare professionals and stakeholders. Ergo, the Figure~\ref{fig:final_agent} presents the operational workflow leveraged in the research. 

Beyond generating predictions, the framework delivers actionable, interpretable insights, bridging the gap between technical model outputs and decision-making processes, while maintaining full transparency and compliance.

\mycomment{
\begin{figure}[ht]
    \centering
    \includegraphics[width=1.1\linewidth]{Agent.png}
    \caption{Agentic AI workflow for healthcare claim analysis. 
    The architecture comprises five specialized sub-agents, - Anomaly Detection, Statistical Insights, Claim Denial Prediction and Explainability, and Root Cause Analysis - enriched by a Hybrid Contextual RAG pipeline and finalized through a HIPAA Compliance Agent.}
    \label{fig:final_agent}
\end{figure}
}

\mycomment{
\begin{figure}[ht]
    \centering
    \includegraphics[width=1.1\linewidth]{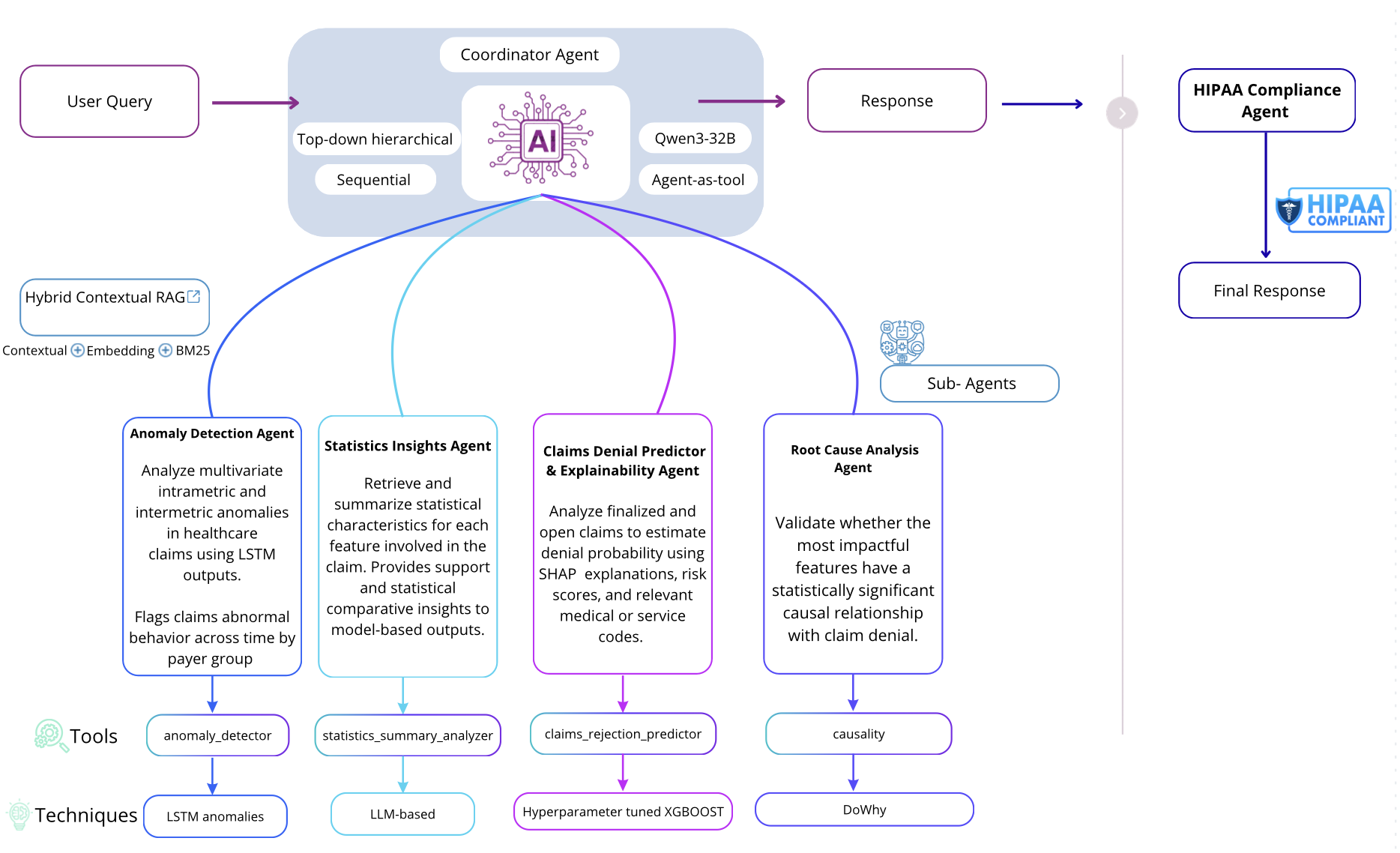}
    \caption{Agentic AI workflow for healthcare claim analysis. 
    The architecture comprises five specialized sub-agents, - Anomaly Detection, Statistical Insights, Claim Denial Prediction and Explainability, and Root Cause Analysis - enriched by a Hybrid Contextual RAG pipeline and finalized through a HIPAA Compliance Agent.}
    \label{fig:final_agent}
\end{figure}
}

\begin{figure}[ht]
    \centering
    \includegraphics[width=1.1\linewidth]{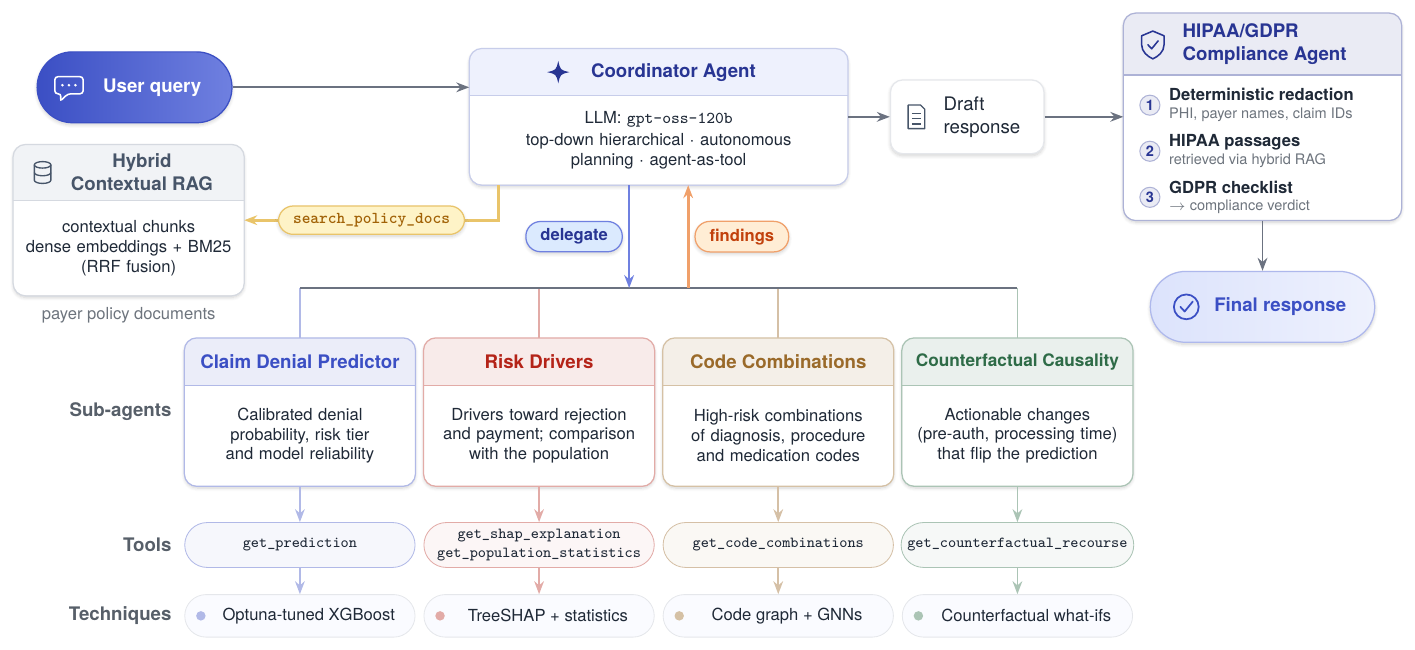}
    \caption{Agentic AI workflow for healthcare claim-denial analysis: a Coordinator Agent delegates to four specialized sub-agents (Claim Denial Predictor, Risk Drivers, Code Combinations, Counterfactual Causality), enriched by a Hybrid Contextual RAG pipeline and finalized by a HIPAA/GDPR Compliance Agent.}
    \label{fig:final_agent}
\end{figure}

The proposed workflow follows a \textbf{top-down hierarchical} architecture using an \textbf{agent-as-tool} pattern. The process is orchestrated by the Coordinator Agent that receives the user query, \textbf{autonomously} selects the specialist agents, retrieves relevant policies, and drafts the report.

Depending on the query type, the coordinator dynamically activates one or more specialized sub-agents, each operating independently with its own prompt pool and reasoning logic, as illustrated.

The draft response is then reviewed by the HIPAA Compliance Agent, while an independent quality gate verifies that all required steps have been completed before release.

\subsection{Agent-Oriented Reasoning \& Tools}

In the reasoning stage, the Coordinator Agent follows a structured reasoning process inspired by CoT \cite{wei2023}, assessing requirements, gathering and reconciling evidence, deriving recommendations, and evaluating confidence based on context quality, relevance, and completeness. Following the \textit{Agent-as-Tool} paradigm \cite{zhang2025agents}, it autonomously selects and invokes specialized sub-agents according to task requirements, avoiding a hard-coded execution sequence. Each sub-agent performs a well-defined analytical function, as summarized in Table~\ref{tab:agents_overview}.

\begin{table}[tbp]
\centering
\caption{Overview of the specialized agents: tools, inputs, and outputs.}
\label{tab:agents_overview}
\small
\setlength{\tabcolsep}{4pt}
\renewcommand{\arraystretch}{1.15}
\begin{tabularx}{\linewidth}{@{}>{\raggedright\arraybackslash}p{0.25\linewidth}>{\raggedright\arraybackslash}X>{\raggedright\arraybackslash}X@{}}
\toprule
\textbf{Agent / Tool} & \textbf{Input} & \textbf{Structured output} \\
\midrule
\textbf{Claim Denial Prediction}
\agenttool{get\_prediction} &
Calibrated denial probability from the Optuna-tuned XGBoost model; test-set reliability metrics; actual outcome (if known) &
Denial probability, predicted outcome vs.\ decision threshold, risk level, and model reliability in plain language \\
\addlinespace
\textbf{Risk Drivers}
\agenttool{get\_shap\_\allowbreak explanation}
\agenttool{get\_population\_\allowbreak statistics} &
Per-claim TreeSHAP contributions as plain-language drivers with strength; claim attributes and population percentiles; payer denial rate &
Drivers toward rejection and toward payment (strong / moderate / small) and comparison with the population \\
\addlinespace
\textbf{Code Combinations}
\agenttool{get\_code\_\allowbreak combinations} &
Code co-occurrence graph with GNN code propensities; pair/triple combination lift; claim-level GNN probability; code descriptions &
High-risk code combinations (labelled codes, denial rate, $\times$ average, support), riskiest single codes, and a HIGH / MEDIUM / LOW verdict \\
\addlinespace
\textbf{Counterfactual Causality}
\agenttool{get\_counterfactual\_\allowbreak recourse} &
Claim re-scored by the calibrated XGBoost model under actionable interventions (approved pre-authorization, typical processing time) &
Probability before $\rightarrow$ after each intervention, decision flips, best achievable outcome, and most effective lever \\
\addlinespace
\textbf{HIPAA / GDPR Compliance} &
Coordinator's draft report; HIPAA passages retrieved from a dedicated index &
Redacted report (PHI, \texttt{PAYER-xxxxxx}, \texttt{CLAIM-xxxxxx}) and a JSON verdict: HIPAA / GDPR compliance, payer anonymity, issues, and citations \\
\bottomrule
\end{tabularx}

\vspace{4pt}
\begin{minipage}{\linewidth}
\footnotesize\raggedright
\textit{Note:} The four analytical tools take \texttt{claim\_id} as their only parameter and return Markdown. The compliance agent receives the claim identifiers for pseudonymization and returns a redacted report plus a JSON verdict.
\end{minipage}
\end{table}
As illustrated, each specialized sub-agent wraps a domain-specific analytical model and exposes it to the Coordinator as a tool. All models are built from the same dataset of 99,079 adjudicated claims (27.4\% denied, where denied means a final status of rejected or partially paid). Thus, sharing the same  stratified train/test split (79,263 / 19,816 claims).

Considering the \textbf{Claim Denial Prediction Agent}, the denial risk is estimated with a gradient-boosted tree model (XGBoost), whose hyper-parameters are tuned via Optuna (TPE sampler, 30 trials, 5-fold stratified cross-validation), and probabilities are calibrated with Platt scaling. The resulting model reaches a ROC-AUC of 0.90 on the held-out test set (precision 0.80, recall 0.76), with configurations details provided in the Table ~\ref{tab:hyperparameters}. For each claim, the agent reports the calibrated denial probability, the predicted outcome against a 0.5 decision threshold, a risk level (low, medium, high, very high), and the model's reliability in plain words.

To explain these predictions, the\textbf{ Risk Drivers Agent }(statistics and explainability) leverages TreeSHAP contributions computed natively by XGBoost for every claim, and population statistics. In this sense, the agent translates the strongest SHAP contributions into plain-language risk drivers, indicating their direction and strength, also contextualizing each claim using population percentiles and payer-specific denial rates, without exposing raw SHAP values.

Beyond individual risk factors, \textbf{GNNs} enable identifying relationships between variables, capturing patterns that may not be apparent when features are analyzed independently. The combination analysis focuses on diagnosis, procedure, and medication codes due to their relevance to healthcare claim adjudication and the potential dependencies between them.

The GNN is a custom edge-conditioned \textbf{Message Passing Neural Network (MPNN)}, inspired by GraphSAGE, with two message-passing layers, a hidden dimension of 64, ReLU activations, and dropout (0.1). 

The model was trained for 200 epochs using Adam (learning rate $0.01$, weight decay $10^{-4}$) and a weighted binary cross-entropy loss to address class imbalance.

Thus, GNNs identify high-risk code pairs and triples, ranking them by denial lift relative to the overall denial rate, while a complementary claim–code graph model provides an additional claim-level risk estimate. The agent reports denial-prone combinations in plain language, including their denial rates and code descriptions, to support interpretation.

Moving beyond risk prediction, the \textbf{Counterfactual Recourse Agent} supports actionable decision-making by estimating how feasible interventions could affect the predicted claim denial probability.

By re-scoring claims  under alternative feature configurations, it performs counterfactual \textit{what-if} analyses to evaluate whether changes to actionable factors, such as, obtaining pre-authorization or reducing processing delays, could alter the predicted outcome. This enables claims analysts and billing teams to explore potential corrective actions, and perform more informed decisions before claim submission or re-submission.

Finally, the \textbf{HIPAA/GDPR Compliance Agent} enforces compliance through three layers: data minimization to prevent unnecessary exposure of PHI, deterministic redaction of sensitive identifiers from reports and execution traces, and an LLM-based review that validates HIPAA compliance and payer anonymity. The reviewer employs retrieved HIPAA guidance and an explicit GDPR checklist, flagging violations and requiring human oversight of final claim decisions. Failed checks trigger a \textit{“do not distribute”} warning and a compliance report documenting the findings and supporting references.

\mycomment{

\subsection{Hybrid Contextual RAG}

The proposed workflow leverages a Hybrid RAG pipeline based on real sources. 

Ergo, the process combines semantic retrieval (vector embeddings) and lexical retrieval (BM25) \cite{anthropic2024}. This dual strategy mitigates the limitations of each method: semantic search captures conceptual relevance but may miss exact term matches, while BM25 \cite{Robertson2009} excels at precise lexical matching but lacks semantic generalization. Together, they balance depth and precision by jointly scoring documents based on semantic similarity, term frequency, and document length. 

Document chunks are further enriched with exploratory context through LLM-based prompting before embedding, situating each segment within its broader document context.

This enhancement overcomes a key limitation of traditional RAG systems, which often treat chunks in isolation and thus lose global coherence \cite{anthropic2024}. At query time, results from both retrieval methods are merged to provide the LLM with a unified, contextually rich evidence base.
}

\subsection{Hybrid Contextual RAG}

The proposed workflow grounds its recommendations in policy evidence through a Hybrid Contextual RAG pipeline \cite{anthropic2024}. Rather than operating as a standalone retrieval stage, the RAG pipeline serves as a shared evidence layer across the workflow, enabling agents to retrieve payer-specific policy information whenever required by their respective tasks.

Considering this, the corpus consists of payer policy documents covering five topics: inpatient hospital services, eligibility and enrollment, skilled nursing facility care, hospice care, and claim denials and appeals, complemented by an industry report on the state of claims. Thus, policy documents were adapted from public health-coverage and policy-research sources, covering multiple fictional payers, aligning policy evidence with the claims data. All documents are explicitly labeled as fictional, not real insurance policies.

Following contextual retrieval \cite{anthropic2024}, documents are divided into chunks of approximately 1,200 characters with 150-character overlap, preserving paragraph and sentence boundaries. Each chunk is enriched with a short, LLM-generated context using \textit{gpt-oss-120b}, identifying its document context, such as the payer and benefit, to improve retrieval coherence and payer-specific queries.

The contextualized chunks are indexed using two complementary methods: BM25 lexical retrieval \cite{Robertson2009}, which captures exact matches for codes and policy terms, and dense semantic embeddings using \textit{BAAI/bge-small-en-v1.5} (384 dimensions, computed locally). This hybrid approach combines semantic relevance with BM25's precise term matching, balancing conceptual understanding and lexical accuracy while accounting for term frequency and document length.

At query time, the top 30 candidates from each method are combined using Reciprocal Rank Fusion \cite{cormack2009rrf}, which merges rankings without requiring score normalization. The five highest-ranked passages are returned with chunk identifiers for inline citation in the final report.

\subsection{Evaluation Metrics}

To evaluate the workflow systematically, we combine \textbf{deterministic} evidence-based metrics computed from the execution trace with an \textbf{agentic self-assessment} produced by the Coordinator.

In this sense, deterministic metrics provide reproducible evidence of observable execution properties, such as policy retrieval, source citation, and compliance with predefined workflow requirements. However, these metrics alone cannot fully capture the contextual adequacy of the provided analysis, so, the Coordinator therefore provides a complementary qualitative assessment of context quality, relevance, and completeness. 

Recent work on evaluating natural-language explanations stresses that trustworthiness requires going beyond text similarity to assess both plausibility and faithfulness \cite{Shailya2025}, and metrics such as groundedness are widely used in agentic and RAG systems to quantify the alignment between a response and its supporting evidence \cite{ragas2025, Ragas}. Following this line, every metric below is computed against the evidence the Coordinator actually received, rather than against a reference text.

The evaluation runs the full workflow, in autonomous mode and with a neutral question that does not name any tool, on a stratified sample of held-out \textbf{high-risk} claims (calibrated denial probability $\geq 0.5$; half in the 0.5--0.8 range and half above 0.8).

\subsubsection{Coordinator Planning}

The proportion of mandatory analysis steps autonomously selected by the Coordinator versus completed by the independent quality gate, based on execution-trace records of each step's trigger.

\subsubsection{Faithfulness}
Faithfulness measures the proportion of numerical values in the report that match tool evidence from specialist agents and retrieved policy passages, within rounding tolerance. All figures, including probabilities, percentages, ratios, amounts, and counts, are checked; model-derived values not explicitly supported by the evidence are considered unsupported, making the metric conservative (0–1).

\subsubsection{Context Quality (Retrieval Grounding)}
Policy context quality is assessed by the number of policy searches per claim and citation validity, measured as the proportion of cited \texttt{[chunk\_id]} passages that exist in the policy index and were retrieved for the query. Invalid or fabricated citations count as failures.

\subsubsection{Privacy and Compliance}
The HIPAA/GDPR Compliance Agent is evaluated by comparing the draft and released reports using deterministic scanners to count payer names, raw claim identifiers, and PHI patterns. The reviewer's HIPAA compliance, GDPR compliance, and payer anonymity verdicts are also recorded for each report.

\subsubsection{Readability}
Readability is assessed by counting technical terms (e.g., SHAP, log-odds, lift, propensity, and ROC-AUC) in released reports and calculating the proportion of reports free of jargon, since the reports are intended for claims and billing staff.

\subsubsection{Model Outcomes and Cost}
For context, we report whether each high-risk claim was actually denied (model correctness on the sample), the share of claims for which the counterfactual recourse finds an actionable change that flips the predicted decision, and the latency and number of tool calls per claim.

\subsubsection{Agentic Self-Assessment}
In addition, the Coordinator is instructed to score its own response in the range $[0,1]$ for \textbf{context quality} (reliability, clarity and consistency of the information used), \textbf{relevance} (pertinence of that information to the task) and \textbf{completeness} (coverage of every aspect of the request). This yields a standardized, response-level confidence signal that is interpretable and auditable, and that can be compared against the deterministic metrics above.

\subsubsection{Statistical Reporting}
All proportions are reported with 95\% Wilson confidence intervals. For metrics with zero observed failures (e.g.\ identifiers in released reports), the upper confidence bound follows the rule of three ($\approx 3/n$).

\section{Data \& Exploratory Data Analysis} \label{sec:data}

Our study relies on a \textbf{synthetic} claim-level dataset that reproduces the structure of a \textbf{U.S.} hospital billing system, namely, claims, payers, clinical codes, pre-authorizations and adjudication outcomes, without containing any real patient or payer information. The generator produced \textbf{100,000 claim} records, of which \textbf{99,079} were retained after applying the eligibility rules described in Table ~\ref{tab:datasets}. 

The generator models dependencies among claim characteristics, administrative processes, clinical codes, and adjudication outcomes through a declarative rule-based approach rather than independent attribute sampling. Claims follow weighted administrative life-cycle templates, with processing times assigned to each transition. 

For claims classified as slow-processing, time intervals are extended by factors of two or three, and delays are sampled from a Beta distribution skewed towards longer durations; otherwise, times are sampled within the normal configured ranges. This mechanism reproduces realistic administrative delays and ageing backlogs. 

Turnaround times are further conditioned on eligibility status, payer, medical specialty, and denial rates, ensuring that administrative delays reflect meaningful relationships with claim outcomes. Finally, targeted anomalies, including missing authorizations, date misalignments, and code mismatches, are injected using independent random seeds to ensure reproducibility and allow new scenarios to be introduced without altering existing data patterns.

Using synthetic data allows the framework, including its privacy and compliance mechanisms, to be developed, evaluated and released openly.

The dataset is summarized in Table~\ref{tab:datasets}, and it was used to evaluate the framework's predictive performance, interpretability, and compliance mechanisms. Preprocessing follows a declarative feature specification.

Skewed monetary variables are Box–Cox transformed, and continuous variables are standardized using z-scores. Categorical variables are one-hot encoded, while the patient's denial ratio over the preceding 120 days and the claim amount per diagnosis are discretized into ordinal bins.

Additional binary indicators capture missing or denied pre-authorisation and a high recent denial ratio. High-cardinality clinical code lists (diagnoses, procedures, and medications) are multi-hot encoded, retaining the 50 most frequent codes per type with a minimum support of 10 claims to limit sparsity. These transformations yield approximately 170 model features.  

A complete summary of the variables, their descriptions, and encoding strategies is provided in Appendix~\ref{app:supplementary} (Table~\ref{tab:appendix_variables}, Table ~\ref{tab:appendix_eda}), 

With 27.4\% denied claims, the target is moderately imbalanced. Rather than resampling, class weighting was treated as a hyper-parameter: the Optuna search could choose \textit{scale\_pos\_weight} between 1 and the negative-to-positive ratio. This approach preserves all training observations and the original feature distribution, avoiding the information loss of undersampling and the duplication or synthetic-sample risks of oversampling.

The selected configuration kept \textit{scale\_pos\_weight} $= 1$, and probabilities were calibrated with Platt scaling, so that the reported denial probabilities remain interpretable.

\begin{table}[t]
\centering
\caption{Overview of the Synthetic Healthcare Claim Denial Dataset.}
\label{tab:datasets}
\small
\renewcommand{\arraystretch}{1.2}
\setlength{\tabcolsep}{5pt}
\begin{tabularx}{\linewidth}{
    >{\raggedright\arraybackslash}p{0.24\linewidth}
    >{\raggedright\arraybackslash}X
}
\toprule
\textbf{Aspect} & \textbf{Description} \\
\midrule

\textbf{Scope} &
Synthetic adjudicated healthcare claims associated with simulated clinical encounters from January to April 2026. \\
\addlinespace[3pt]

\textbf{Volume} &
99,079 eligible claims retained from 100,000 generated records containing 81 columns. Eligibility requires a final claim status and a billed amount greater than \$10. Each record is identified by a \texttt{claim\_id}. \\
\addlinespace[3pt]

\textbf{Target Variable} &
Binary classification: \newline
Denied/Partially Paid (Class 1): 27,183 claims (27.4\%). \newline
Paid (Class 0): 71,896 claims (72.6\%). \\

\addlinespace[3pt]
\textbf{Data Realism} &
Real coding systems, including ICD-10 diagnosis codes, procedure, and medication codes with descriptions. Payer names, patient history and identifiers, are fictional. \\

\addlinespace[3pt]
\textbf{Feature Composition} &
Claim attributes (amount, type, setting, month, and day of week); administrative variables (processing time and pre-authorization status); payer and patient history (average amounts, claim counts, and recent denial ratio); demographics (age band and gender); and clinical codes (diagnoses, procedures, and medications), comprising 143 distinct codes. \\
\addlinespace[3pt]
\textbf{Data Split} &
A single stratified 80/20 train--test split, comprising 79,263 training claims and 19,816 test claims, shared across all evaluated models. \\

\bottomrule
\end{tabularx}
\end{table}

\section{Results}\label{sec:experiments}

This section presents and discusses the main findings from the experimental pipeline. The results are structured across different framework configurations to demonstrate how each component contributes to answering the RQ, namely through evaluating baseline reasoning, and compliance validation with HIPAA safeguards.

In this regard, Hybrid Contextual RAG is incorporated into every configuration, ensuring that the coordinator and its selected specialist agents consistently ground their recommendations in the retrieved payer-policy documents. The results therefore compare two framework configurations that differ only in the application of the compliance layer:

\mycomment{
\begin{enumerate}
    \item \textbf{Standard configuration:}
    The draft report produced by the coordinator after orchestrating the specialist agents and retrieving relevant payer-policy documents
    through contextual RAG, without HIPAA safeguards.

    \item \textbf{HIPAA-compliant configuration:}
    The same report after review by the HIPAA/GDPR compliance agent. This agent redacts protected health information (PHI), replaces payer names with pseudonyms, and provides a compliance verdict grounded in the retrieved HIPAA guidance.
\end{enumerate}
}

\subsection{Standard Framework}
To test our hypothesis, we first implemented a baseline version of the proposed Agentic AI framework without the HIPAA Compliance Agent, while retaining hybrid contextual RAG to ensure that the coordinator and specialist agents grounded their recommendations in the retrieved payer-policy documents. 

This experimental strategy aimed to assess the core reasoning capabilities of the Coordinator Agent within a hybrid contextual RAG setting, enabling an in-depth investigation of its inference performance before introducing the additional HIPAA compliance safeguards.

The following illustration~\ref{fig:case_study} presents an illustrative case study involving a held-out, high-risk claim from the synthetic dataset (claim 604478). The user requests an assessment of the claim's denial risk and its main drivers, an analysis of high-risk code combinations, potential actions to mitigate the risk, and supporting evidence from the payer's policy documents. 

Following the structured reasoning procedure described in Section~\ref{sec:methodology} (assess, plan and gather evidence, reconcile, recommend, and score confidence), the Coordinator orchestrated the four specialist agents and queried the policy knowledge base five times before synthesizing the final report. 

\clearpage

\mycomment{
\foreach \i in {1,...,4} {
    \includepdf[pages=-, scale=0.9, pagecommand={\thispagestyle{fancy}}]{case_study_p\i.pdf}}
}

\includepdf[pages=-, scale=0.90, offset=0 -0.5cm, pagecommand={\thispagestyle{fancy}\captionof{figure}{Standard Framework: high-risk claim, showing the Coordinator's orchestration, agent outputs, and policy-grounded recommendations.}\label{fig:case_study}}]{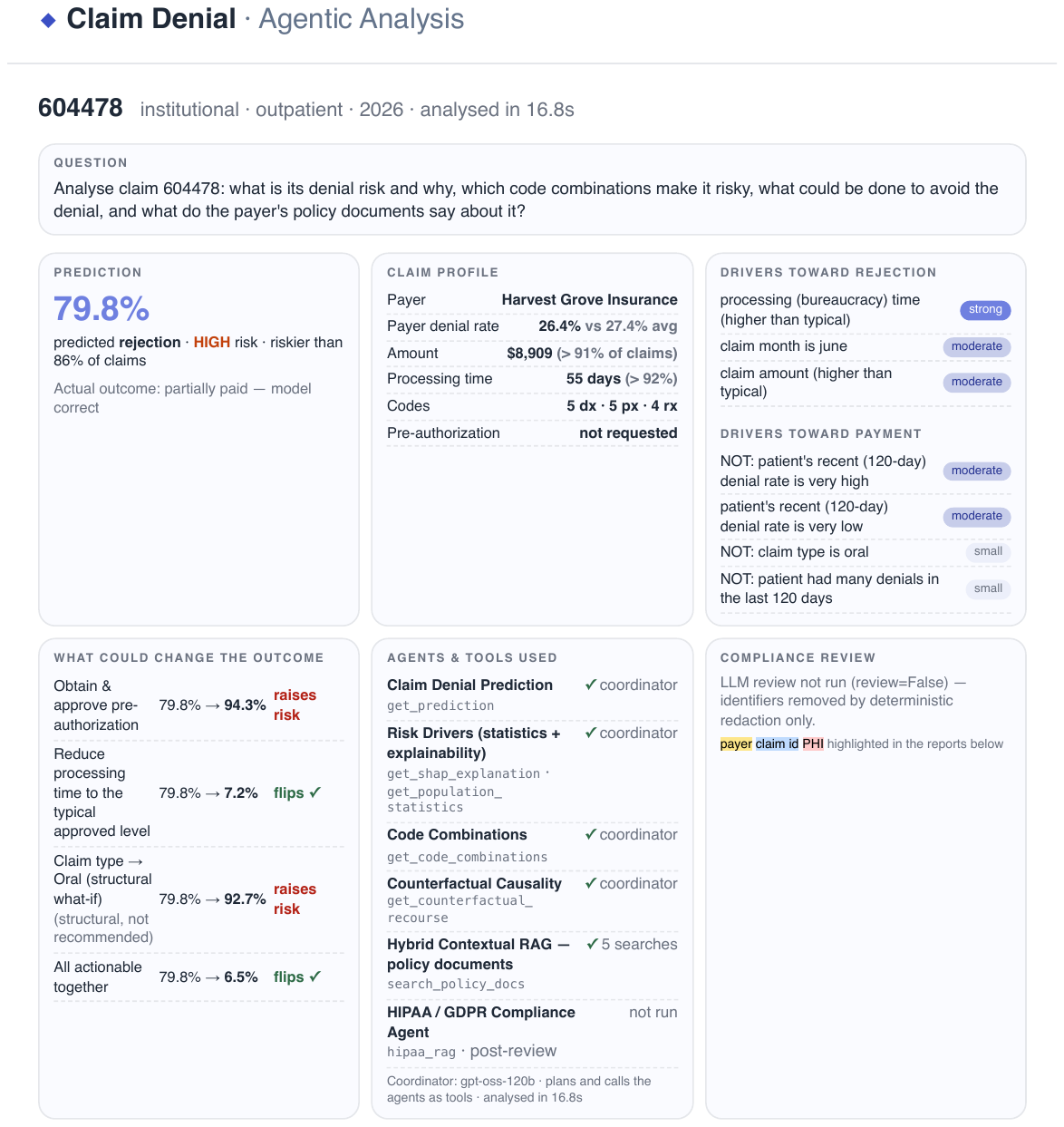}
\includepdf[pages=-, trim=1cm 1cm 1cm 1cm, clip, width=\textwidth, height=0.85\textheight, keepaspectratio, offset=0 2.5cm, pagecommand={\thispagestyle{fancy}\addtocounter{figure}{-1}\captionof{figure}[]{Standard Framework (continued).}}]{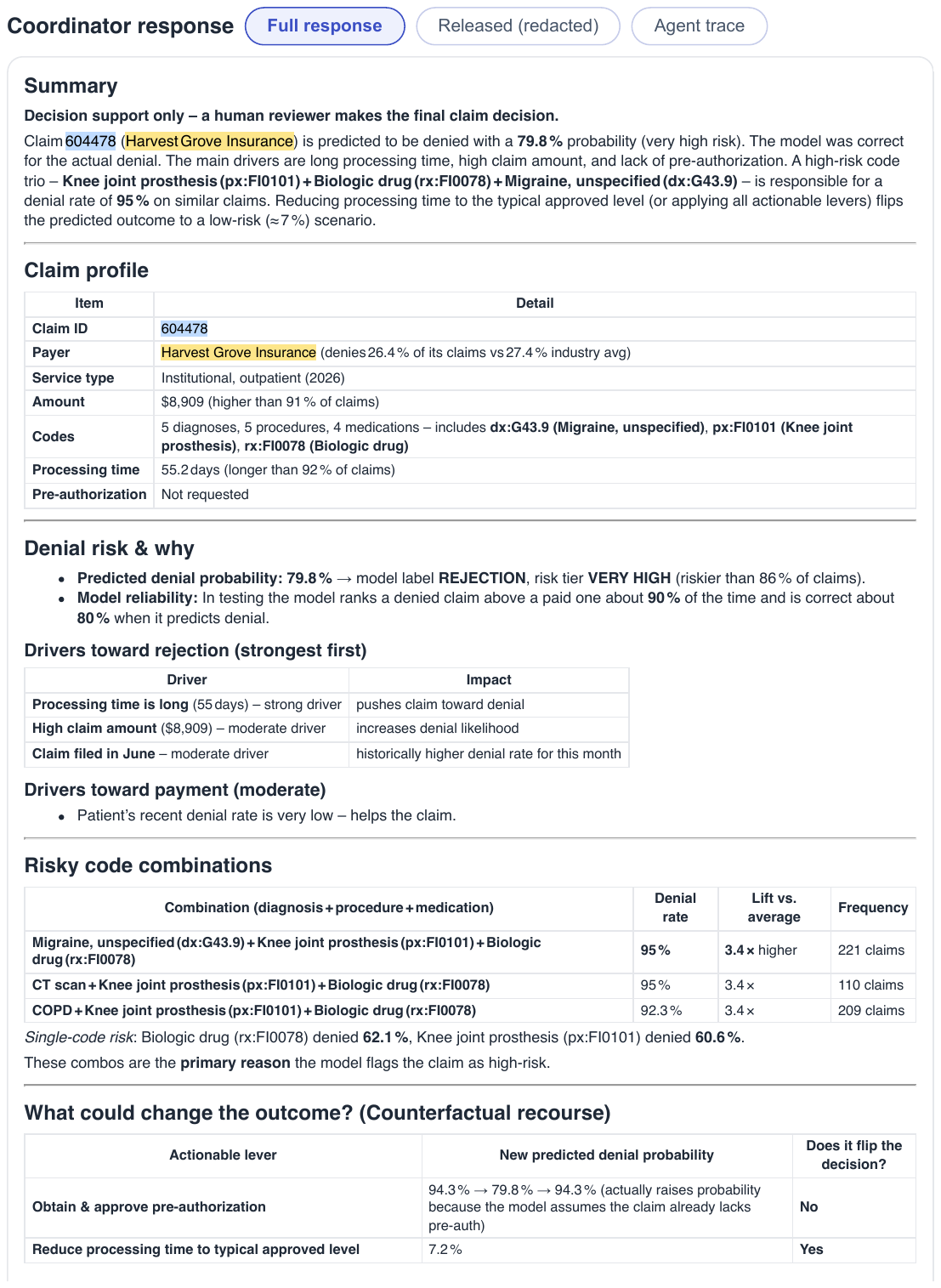}
\includepdf[pages=-, trim=1cm 1cm 1cm 1cm, clip, width=\textwidth, height=0.85\textheight, keepaspectratio, offset=0 2.5cm, pagecommand={\thispagestyle{fancy}\addtocounter{figure}{-1}\captionof{figure}[]{Standard Framework (continued).}}]{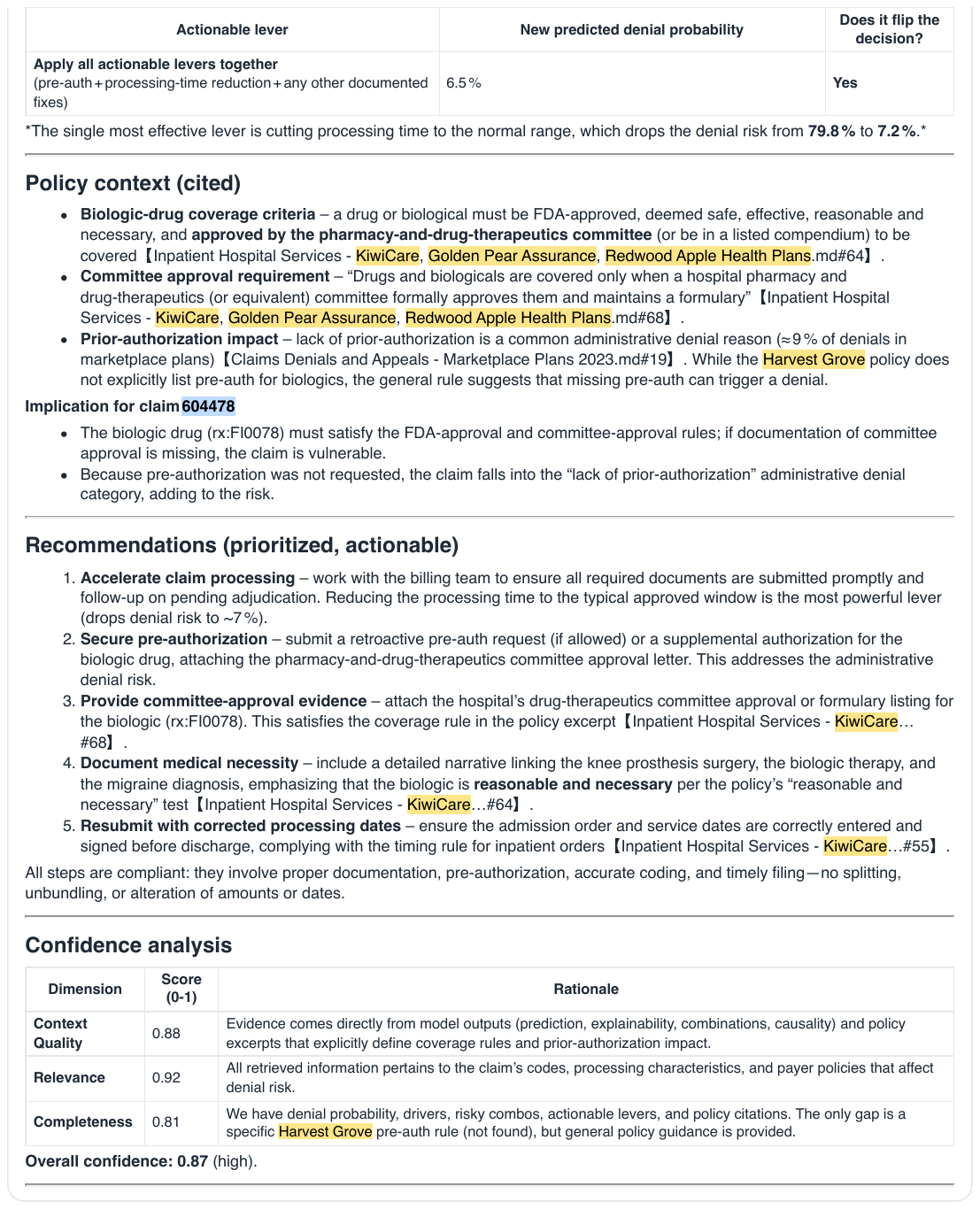}
\includepdf[pages=-, trim=1cm 1cm 1cm 1cm, clip, width=\textwidth, height=0.85\textheight, keepaspectratio, offset=0 2.5cm, pagecommand={\thispagestyle{fancy}\addtocounter{figure}{-1}\captionof{figure}[]{Standard Framework (continued).}}]{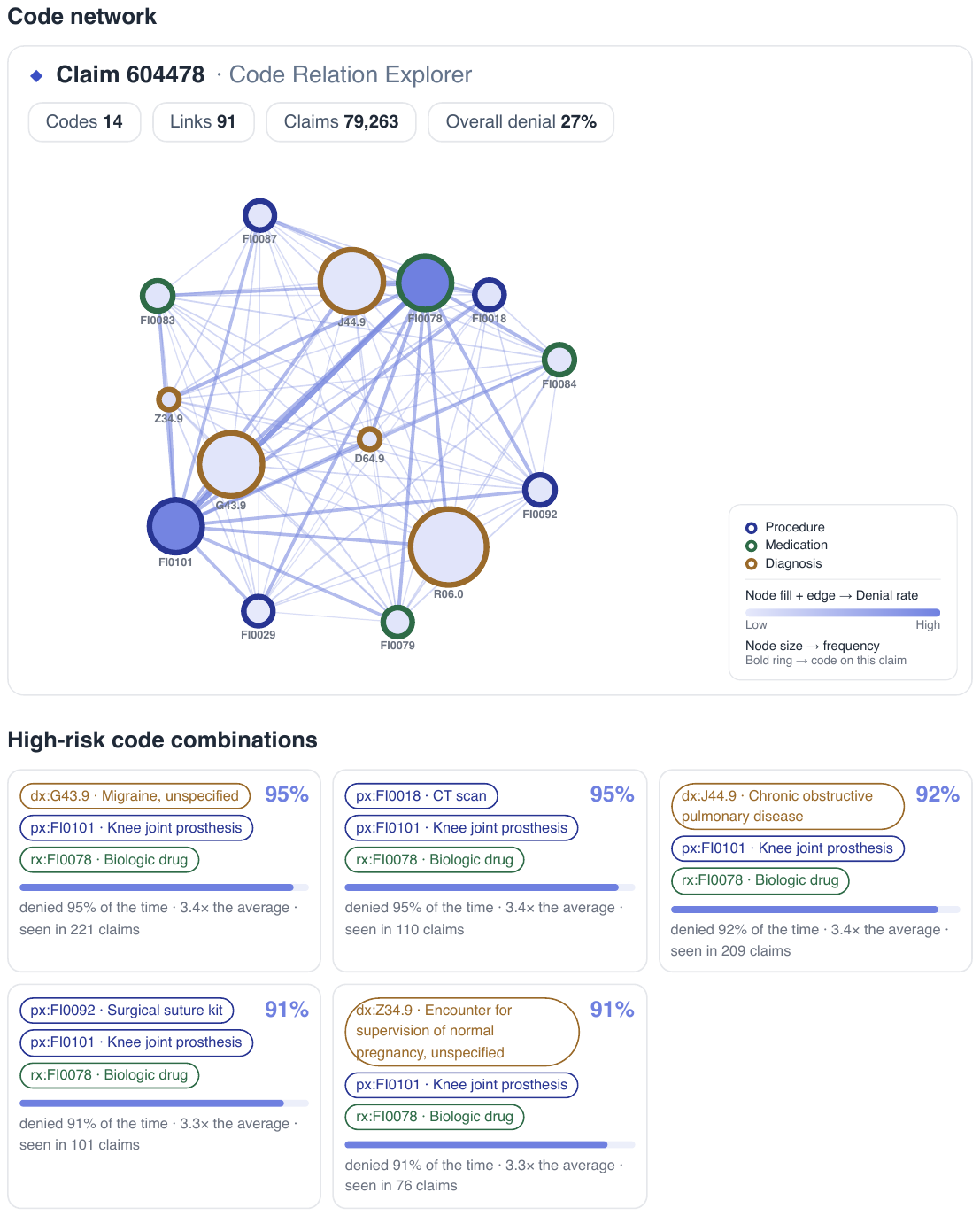}

The results enhanced that the Coordinator Agent demonstrated robust autonomous reasoning and effective tool selection. Given a single question about claim 604478, it planned the analysis from the tool descriptions alone, without being told which tools to use, and called every specialist the question required.

To answer the request, the Coordinator activated 14 tool calls across 10 distinct tools:four specialist agents once each and searched the policy documents five times, and the specialists made five calls to their data tools.

\begin{itemize}
    \setlength{\itemsep}{5pt}
    \item \texttt{run\_prediction\_agent} estimated a denial probability of 79.8\%, classifying the claim as high risk; the claim was ultimately partially paid.

    \item \texttt{run\_explainability\_agent} identified processing time (55 days, exceeding 92\% of claims) and claim amount (\$8{,}909) as the main risk drivers, contextualising the payer's denial rate (26.4\%) against the average (27.4\%).

    \item \texttt{run\_combinations\_agent} identified high-risk code combinations with denial rates of up to 95\%, or 3.4 times the average. The riskiest combinations involved a knee joint prosthesis and a biologic drug paired with migraine (95\% denial rate; 221 claims), a CT scan (95\%; 110 claims), or chronic obstructive pulmonary disease (92\%; 209 claims).

    \item \texttt{run\_causality\_agent} found that reducing processing time to the typical level observed for paid claims lowered the predicted denial probability to 7.2\%, whereas obtaining prior authorization alone was insufficient to achieve a comparable reduction.
\end{itemize}

Finally, the Coordinator queried the payer's policy documents five times through \texttt{search\_policy\_docs} to ground its recommendations, completing the full analysis in about \textbf{17 seconds}.

In this sense, the pipeline identifies the main factors associated with a claim's denial risk and recommends potential next steps, helping teams prioritize high-risk claims and support denial prevention.

Nevertheless, this configuration revealed two limitations. First, although the Coordinator grounded its recommendations in retrieved payer-policy documents, the knowledge base contained no payer-specific pre-authorization rule for the services involved. This gap was reflected in its self-assessed confidence scores: 0.88 for context quality, 0.92 for relevance, and 0.81 for completeness, yielding an overall confidence of 0.87. The reliance on general policy guidance limited the ability to validate payer-specific coverage requirements and reference applicable clinical guidelines.

Second, the configuration operated without the HIPAA Compliance Agent, leaving the raw claim identifier and payer name exposed in the generated report. Such disclosure would be inappropriate in a production environment. These limitations highlight the need to combine policy-grounded reasoning with compliance verification to improve the reliability and safe deployment of the framework.

\subsection{HIPAA Compliance Framework}

To improve output reliability and support regulatory compliance, the proposed workflow extends the conventional standard RAG architecture with a dedicated \textbf{HIPAA  Compliance Agent}. Operating downstream of the Coordinator, this agent validates the completed report without influencing the preceding analysis, comprising two steps:

\begin{enumerate}
    \setlength{\itemsep}{5pt}
    \item \textbf{Deterministic redaction.}
    Rule-based scanners detect and remove patterns associated with PHI, including Social Security numbers, phone numbers, full dates, and record numbers. Payer names and claim identifiers are replaced with stable pseudonyms (\texttt{PAYER-xxxxxx} and \texttt{CLAIM-xxxxxx}).

    \item \textbf{LLM-based compliance review.}
    Grounded in retrieved guidance summarising the HIPAA Privacy Rule, the reviewer assesses the redacted report against the minimum-necessary standard and relevant GDPR principles, including data minimization, pseudonymization, and safeguards concerning solely automated decision-making (Article~22). It returns a structured verdict identifying potential compliance issues and providing supporting citations.
\end{enumerate}

Hence, to evaluate its effect, the Compliance Agent was applied to the response generated for the case study above,
without re-running the analysis, so the content of the report is unchanged and only its privacy handling differs.

In the HIPAA-reviewed version, the claim identifier and the payer name are replaced by stable pseudonyms throughout the text (\texttt{CLAIM-5C7F41},
\texttt{PAYER-E3F1E8}), four payer-name occurrences were redacted, and the verdict confirms that the report is
HIPAA and GDPR-compliant and that no payer can be identified. 

Beyond redaction, the agent appends a compliance
section to the report (Figure~\ref{fig:case_study_reviewed}): the verdict and redactions applied, handling guidance stating
who may use the report and for what purpose, and the passages of the HIPAA Privacy Rule summary that support them.

\mycomment{
\includepdf[pages=-, scale=0.90, offset=0 -0.5cm, pagecommand={\thispagestyle{fancy}\captionof{figure}{HIPAA Compliance Framework: compliance verdict, redactions applied, handling guidance for the report, and the HIPAA Privacy Rule.}\label{fig:case_study_reviewed}}]{case_study_reviewed_p1.pdf}
\includepdf[pages=-, trim=1cm 1cm 1cm 6cm, clip, width=\textwidth, height=0.85\textheight, keepaspectratio, offset=0 -0.5cm, pagecommand={\thispagestyle{fancy}\addtocounter{figure}{-1}\captionof{figure}[]{HIPAA Compliance Framework (continued)}}]{case_study_reviewed_p2.pdf}
\includepdf[pages=-, trim=1cm 1cm 1cm 6cm, clip, width=\textwidth, height=0.85\textheight, keepaspectratio, offset=0 -0.5cm, pagecommand={\thispagestyle{fancy}\addtocounter{figure}{-1}\captionof{figure}[]{HIPAA Compliance Framework (continued)}}]{case_study_reviewed_p3.pdf}
\includepdf[pages=-, trim=1cm 1cm 1cm 6cm, clip, width=\textwidth, height=0.85\textheight, keepaspectratio, offset=0 -0.5cm, pagecommand={\thispagestyle{fancy}\addtocounter{figure}{-1}\captionof{figure}[]{HIPAA Compliance Framework (continued)}}]{case_study_reviewed_p4.pdf}
}

\includepdf[pages=-, scale=0.90, offset=0 -0.5cm, pagecommand={\thispagestyle{fancy}\captionof{figure}{HIPAA Compliance Framework: compliance verdict, redactions applied, handling guidance for the report, and the HIPAA Privacy Rule.}\label{fig:case_study_reviewed}}]{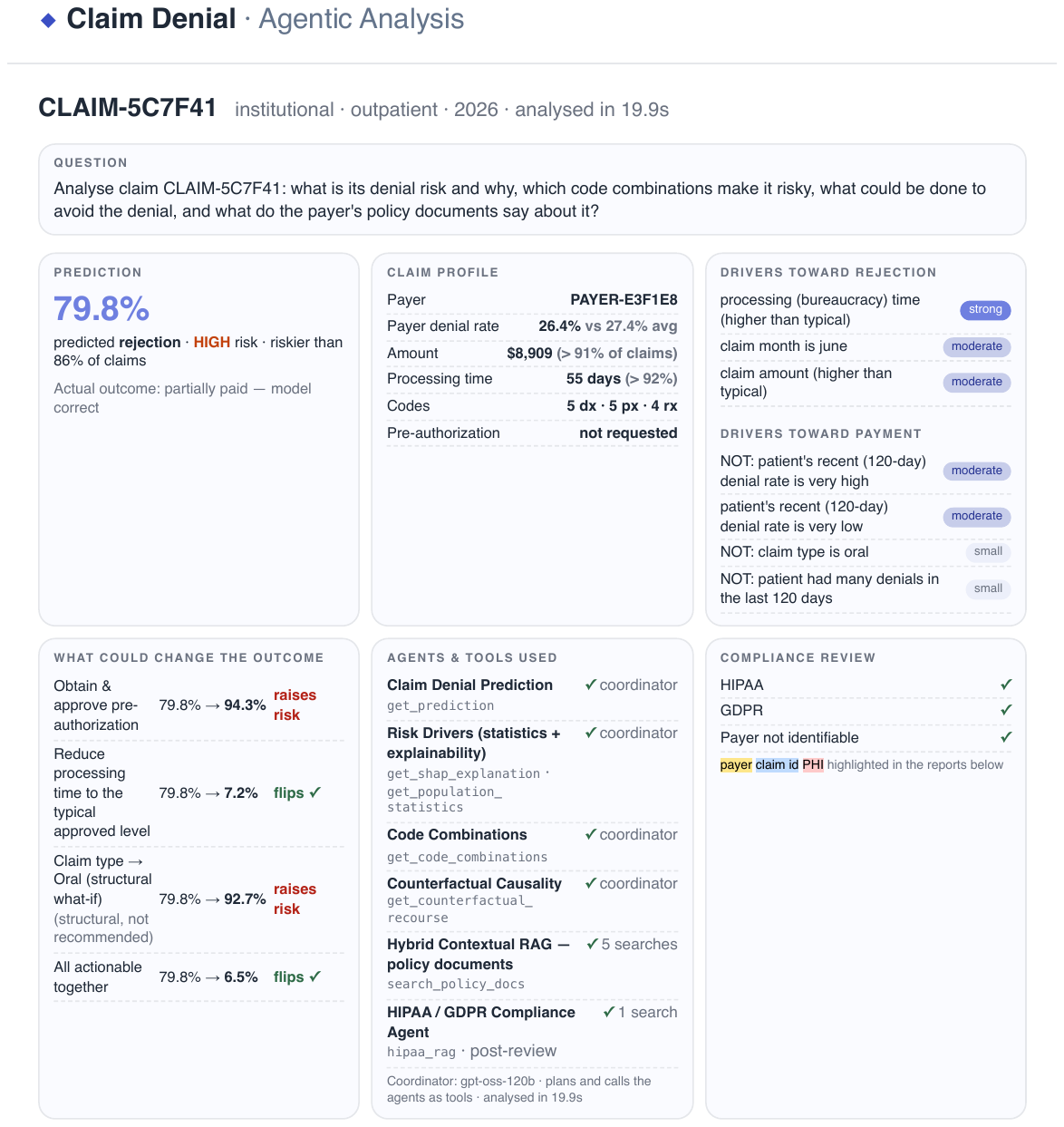}
\includepdf[pages=-, trim=1cm 1cm 1cm 1cm, clip, width=\textwidth, height=0.85\textheight, keepaspectratio, offset=0 2.5cm, pagecommand={\thispagestyle{fancy}\addtocounter{figure}{-1}\captionof{figure}[]{HIPAA Compliance Framework (continued).}}]{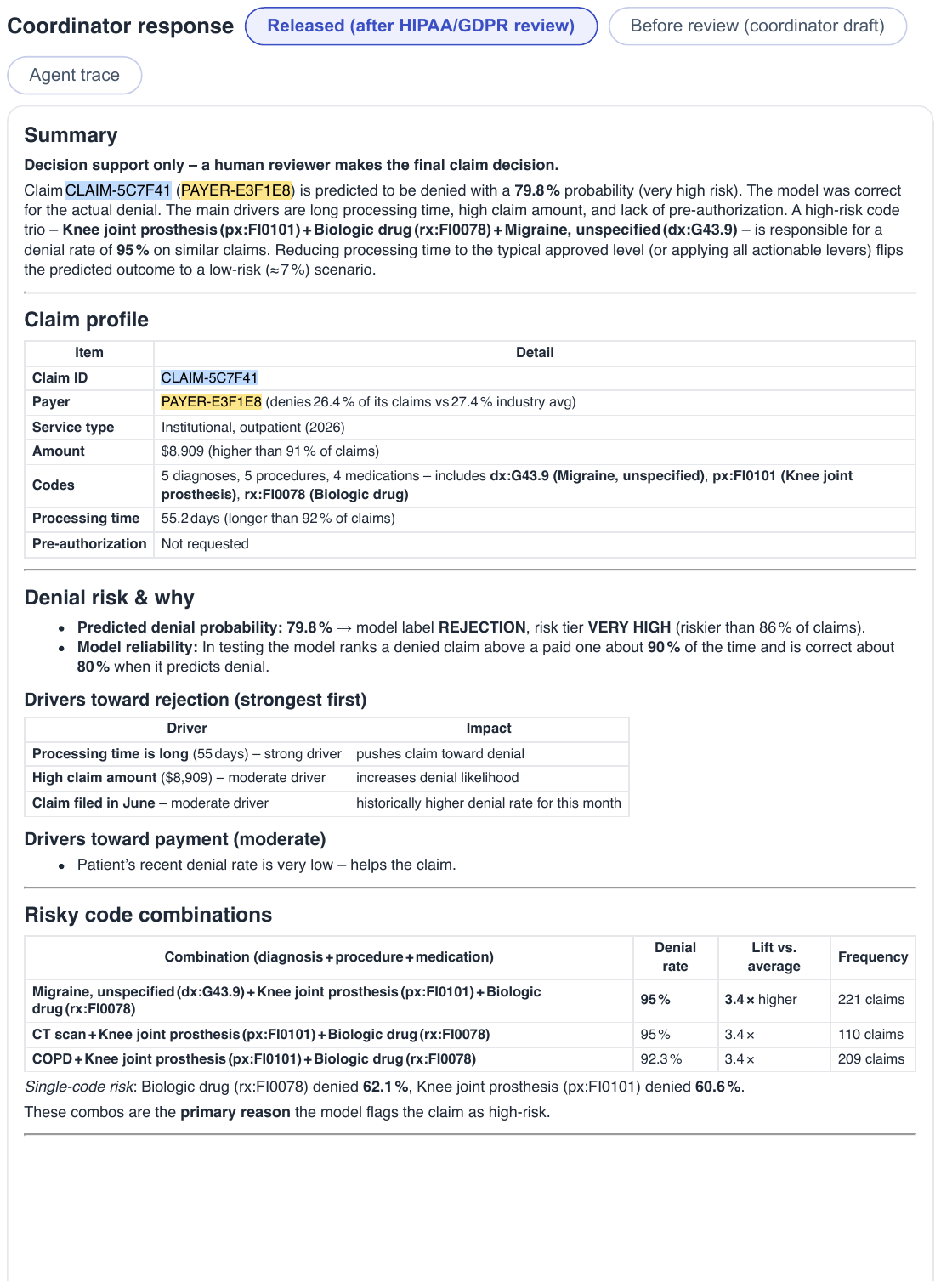}
\includepdf[pages=-, trim=1cm 1cm 1cm 1cm, clip, width=\textwidth, height=0.85\textheight, keepaspectratio, offset=0 2.5cm, pagecommand={\thispagestyle{fancy}\addtocounter{figure}{-1}\captionof{figure}[]{HIPAA Compliance Framework (continued).}}]{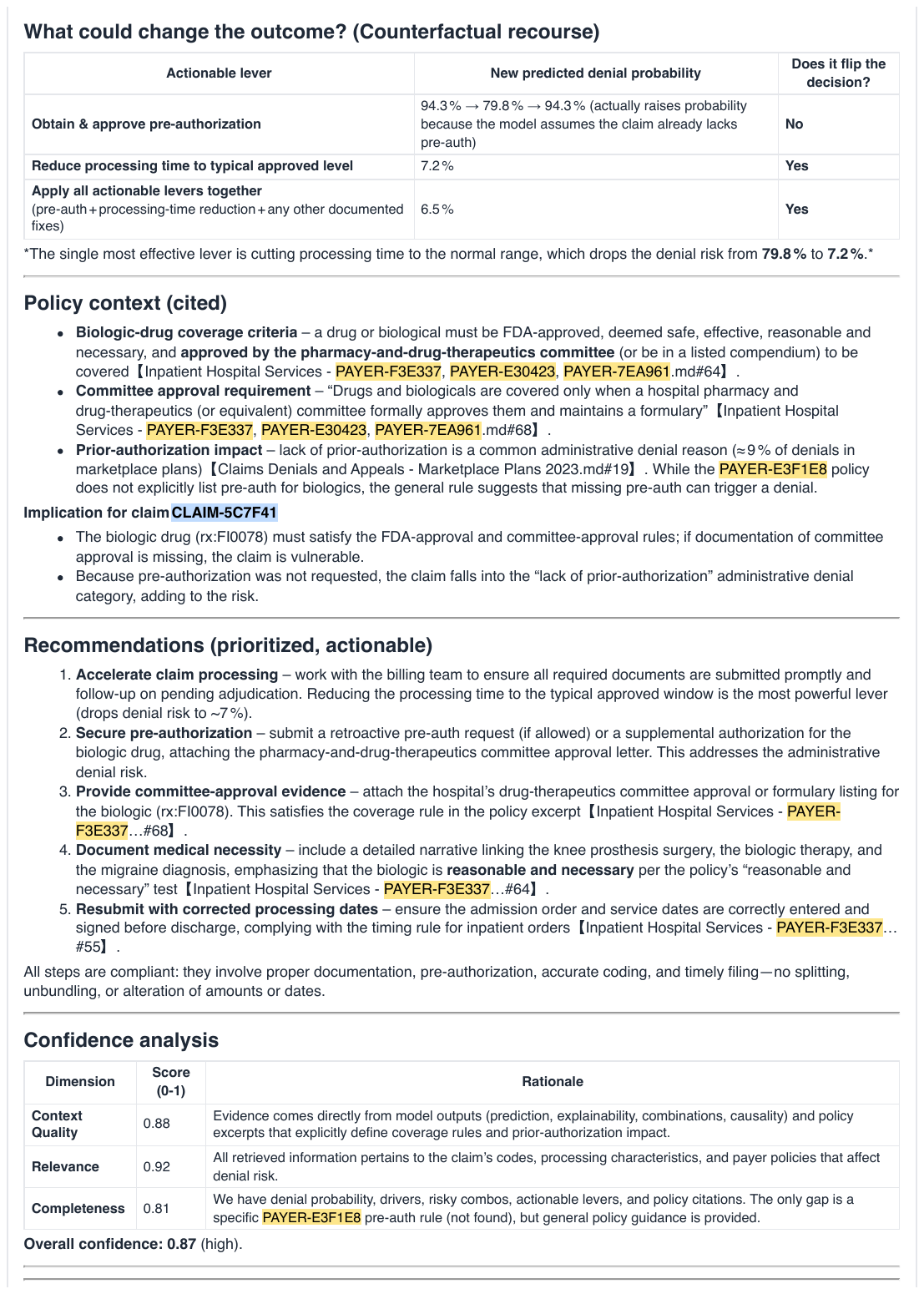}
\includepdf[pages=-, trim=1cm 1cm 1cm 1cm, clip, width=\textwidth, height=0.85\textheight, keepaspectratio, offset=0 2.5cm, pagecommand={\thispagestyle{fancy}\addtocounter{figure}{-1}\captionof{figure}[]{HIPAA Compliance Framework (continued).}}]{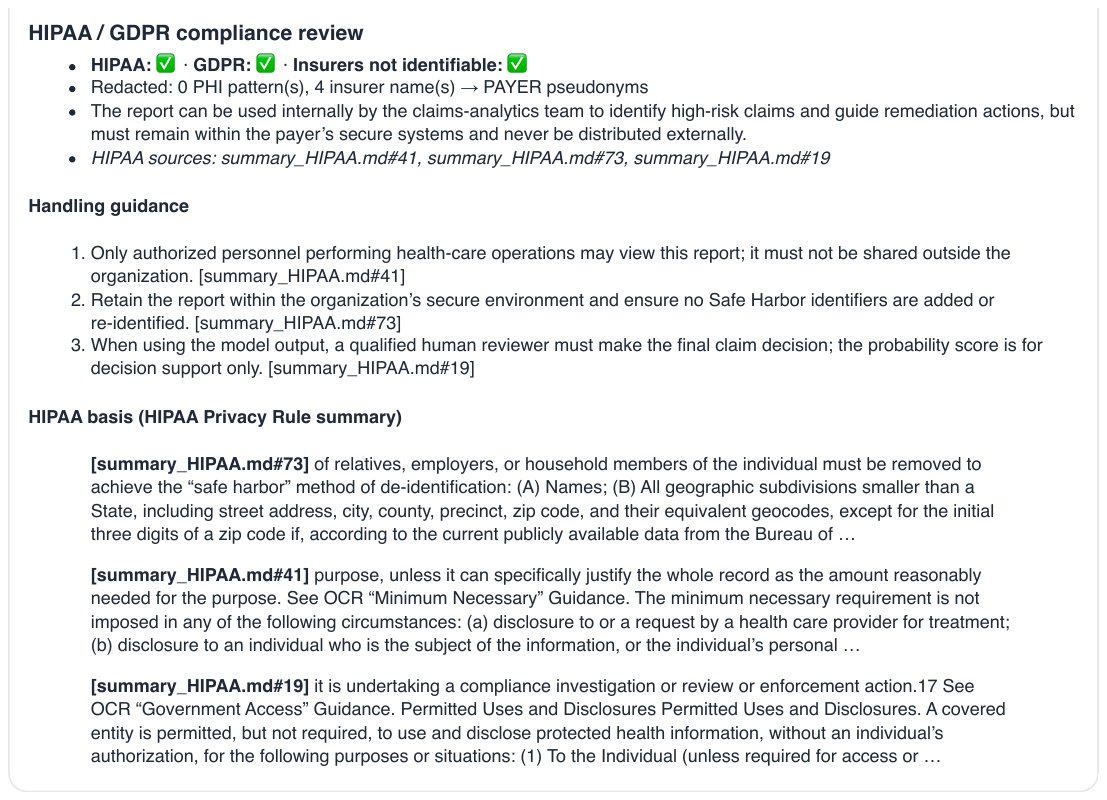}
\includepdf[pages=-, trim=1cm 1cm 1cm 1cm, clip, width=\textwidth, height=0.85\textheight, keepaspectratio, offset=0 2.5cm, pagecommand={\thispagestyle{fancy}\addtocounter{figure}{-1}\captionof{figure}[]{HIPAA Compliance Framework (continued).}}]{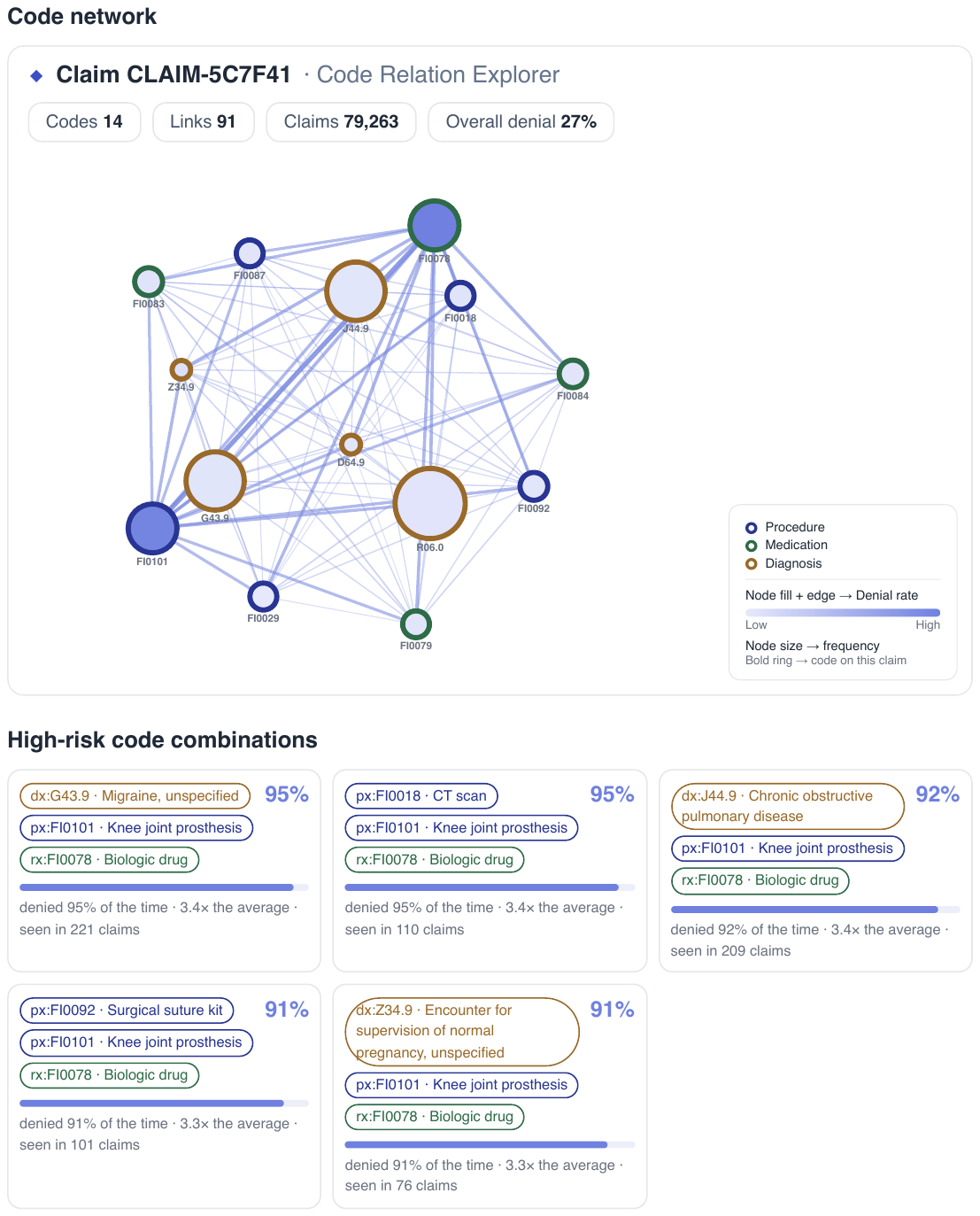}

\mycomment{
From this standpoint, the HIPAA Compliance Agent identified and categorized the disclosure of medication codes as Low-risk, since medication codes solely do not constitute PHI, unless in episodes linked to identifiable individuals.  Given the fact that the payer group names were anonymized from the outset, they are non-identifiable, and do not pose any privacy risk.

Subsequently, the Compliance Agent HIPAA sent this feedback to the Coordinator Agent, which addressed the identified non-compliance subjects, namely the claim identifier, by replacing it with a generic placeholder (\textit{redacted\_claim\_id}). This approach effectively removes sensitive information while preserving the analytical integrity of the response. 

Notably, the remainder of the response is preserved, with only sensitive elements modified, ensuring that the quality produced by the pipeline remains intact, according to pre-defined criteria, mitigating the requirement for manual examination. 

Therefore, this integration enhances the proposed Agentic AI healthcare legal compliance, mitigating the risk of unauthorized PHI exposure. This outcome is reinforced by all evaluated quality metrics yielding the maximum possible score (1.0), denoting optimal performance across every assessment dimension.
} 

From this standpoint, the Compliance Agent detected and pseudonymized the claim identifier and four occurrences of payer names, while finding no other PHI patterns. The LLM reviewer subsequently assessed the report against HIPAA and GDPR requirements, finding it compliant across all three checks (HIPAA, GDPR, and payer anonymity), also providing guidance on internal use, identifier protection, and human oversight.

Importantly, the agent preserved the analytical content, including the denial probability, risk drivers, code combinations, counterfactual scenarios, and policy citations. The Coordinator's evidence confidence scores also remained unchanged (context quality: 0.88, relevance: 0.92, completeness: 0.81), demonstrating that identifier protection can be applied without altering the report's decision-support content.

\subsection{Framework Performance Evaluation}

The comparative analysis was designed to empirically assess the contribution of the HIPAA Compliance Agent to the framework's overall performance. 

In particular, it examines whether the compliance layer improves the protection of sensitive information and adherence to relevant regulatory requirements while preserving the analytical content and decision-support value of the generated reports. 

The framework was evaluated on a stratified sample of \textbf{100 }held-out \textbf{high-risk claims} (50 with a denial probability
between 50\% and 80\%, 50 above 80\%), drawn from the 5,248 high-risk claims in the test set.

As a sanity check, each claim received the same neutral question, which names no tool, and the workflow ran in autonomous mode.

Given the fact that the Coordinator's draft
is stored before the compliance review, each run yields two configurations of the same analysis, namely, without and with the Compliance Agent, so their comparison isolates the effect of the compliance layer.

Thus, by comparing both configurations on the same claims, the evaluation isolates the effect of the compliance layer, providing evidence of its added value and identifying potential trade-offs between regulatory safeguards and analytical utility.

Considering this, Figure~\ref{fig:eval} summarizes the results; proportions are reported with 95\% Wilson confidence intervals, with more details in Appendix Table ~\ref{tab:appendix_evaluation}.

\clearpage

\begin{center}
    \captionof{figure}{Evaluation results on 100 held-out high-risk claims.}
    \label{fig:eval}
    \includegraphics[width=\linewidth, height=0.9\textheight, keepaspectratio,
                     trim=1cm 2cm 1cm 2cm, clip]{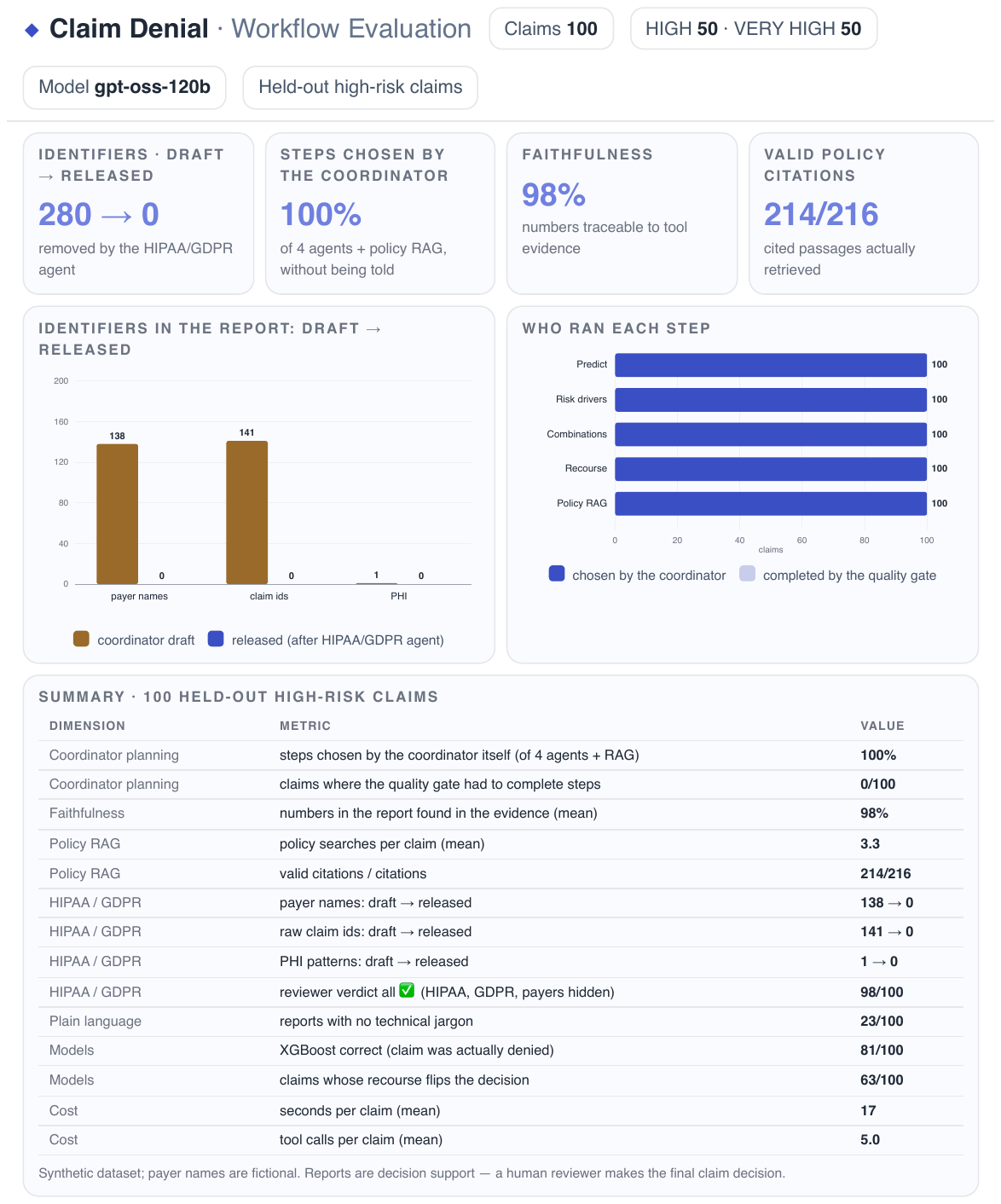}
\end{center}

\textbf{Coordinator planning.} Without being told which tools to use, the Coordinator selected all four specialist
agents and the policy retrieval in every case (100\%, 95\% CI 96.3--100\%), so the independent quality gate never had
to intervene. In addition, it searched the policy documents 3.3 times per claim on average.

\textbf{Quality gate.} Coverage was guaranteed by an
independent quality gate that runs after the Coordinator answers. Using the execution trace, it checked that every claim analysis rests on the four specialist agents and at least one policy search. 

If a step is missing, the gate runs it itself and hands the results back to the Coordinator as \emph{additional evidence}, without naming any tool, and the
Coordinator then rewrites its answer. Every step in the trace records who triggered it (Coordinator or quality gate),
which makes the Coordinator's autonomy measurable rather than assumed.

\textbf{Faithfulness and grounding.} On average, 98\% of the numbers in each report were traceable to the evidence
returned by the tools. Of 216 policy citations, 214 referred to passages actually retrieved for that question (99\%,
95\% CI 96.7--99.7\%); none of the remaining two pointed to a fabricated document.

\textbf{Privacy and compliance.} Without the compliance layer, the drafts exposed 280 identifiers across the 100
reports: 138 payer names, 141 raw claim identifiers and one PHI pattern. In the released reports, none remained.

By the rule of three, this bounds the rate of reports leaking an identifier at $\leq 3\%$ with 95\% confidence.

The LLM reviewer judged 98 of the 100 released reports fully compliant (HIPAA, GDPR and payer anonymity; 95\% CI
93.0--99.4\%). 

\textbf{Models.} The calibrated XGBoost model was correct for 81 of the 100 high-risk claims, which were in fact denied
(95\% CI 72.2--87.5\%), consistent with its test precision of 0.80. 

The counterfactual recourse found an actionable
change that reversed the predicted denial for 63 claims (95\% CI 53.2--71.8\%), most often bringing the processing time
to the typical level of paid claims.

\textbf{Plain language.} Readability was the weakest dimension: only 23 of the 100 reports were free of technical
jargon (95\% CI 15.8--32.2\%), the residual term being almost always ``lift''. Style instructions in the prompt proved
less reliable than the rules governing data and identifiers, which suggests that, like redaction, terminology is better
enforced deterministically.

\textbf{Cost.} A complete analysis took 17 seconds per claim on average, with five data-tool calls by the specialists.

\section{Conclusions} \label{sec:conclusions}

In an era of increasing regulatory scrutiny, the widespread adoption of AI systems depends on frameworks that reconcile legal obligations with practical business requirements.

This research presented an Agentic AI framework for structured, explainable, and policy-aware healthcare claim analysis. Its main contribution lies in integrating specialist agents grounded in machine-learning and graph neural network outputs with a Hybrid Contextual RAG pipeline over payer policy documents and a dedicated HIPAA/GDPR Compliance Agent, providing an integrated decision-support approach that combines predictive performance, interpretability, policy-grounded reasoning, and privacy safeguards while preserving human oversight.

Additionally, the evaluation on 100 held-out high-risk claims showed that the Coordinator autonomously planned the analysis in every case, and the compliance layer removed all the identifiers detected in the draft reports without modifying their analytical content.

These findings address the research question by demonstrating how grounding language-model reasoning in model outputs and retrieved policy evidence, while performing compliance verification downstream, can support healthcare claim assessment under explicit privacy safeguards. 

Beyond the technical evaluation, previous studies have reported substantial operational benefits from AI-driven claim management, including a 50\% reduction in processing time, a 133\% increase in employee productivity, and a 44\% reduction in operational costs, alongside improvements in fraud detection and customer satisfaction \cite{Keshetti2025}. Similarly, multi-agent AI frameworks have been associated with efficiency gains of 40--60\% across financial and operational workflows \cite{Joshi2025}. Although these findings provide a broader motivation for automation, the present study did not directly measure such business outcomes; their achievement within the proposed framework therefore remains to be established empirically.

The framework also addresses ethical, privacy, and compliance risks identified as significant vulnerabilities of agentic AI \cite{gridach2025}. It combines model-grounded reasoning, curated policy evidence, deterministic identifier redaction, and a dedicated compliance review. Restricting retrieval to curated policy documents further reduces exposure to unverified external information, although it also makes the framework dependent on the coverage and currency of its knowledge base.

Ultimately, this research contributes to the development of more trustworthy agentic AI systems by demonstrating an approach that combines predictive analytics, policy-grounded reasoning, and privacy safeguards within a human-supervised decision-support framework. Its practical value will depend on further validation in real-world settings, where analytical reliability, regulatory requirements, and operational needs must be addressed together.

\subsection{Limitations} 

Several limitations must nevertheless be acknowledged. Compliance with stylistic requirements remained inconsistent, with only a small fraction of the reports avoiding technical jargon entirely. Hence, the compliance agent supports risk detection and mitigation but does not, by itself, establish legal compliance or replace human oversight. Deployment through a cloud inference API would also require appropriate contractual, security, and data-protection arrangements. 
In addition, the evaluation covered only 100 high-risk claims from a single synthetic dataset, balancing statistical precision against inference cost, where further validation on larger samples and real-world claims data is necessary.

\subsection{Future Work} 
At the architectural, further research should investigate  \textit{Agent-to-Agent (A2A)} communication and the \textit{Model Context Protocol (MCP) }could improve interoperability and data integration. Finally, on-premises deployment of open-weight models could help organizations retain greater control over sensitive data, subject to appropriate security and governance measures.

\mycomment{
\bmhead{Acknowledgements}
This article is a result of the project CDMS - Claim Denial Management Solution, supported by Innovation and Digital Transition Thematic Programme (COMPETE 2030) and Lisboa Regional Operational Programme (LISBOA 2030), under the PORTUGAL 2030 Partnership Agreement, through the European Regional Development Fund (ERDF).

\bmhead{Author Contributions} 
All authors contributed, read, and approved this manuscript.

\section*{Declarations}
\bmhead{Institutional Conflicts} Randy Labs, Effy Business, School of Economics and Management (University of Porto).  

\bmhead{Individual Conflicts} Ricardo Macedo  (INESC TEC Researcher), João Tiago Paulo (INESC TEC Researcher).

\bmhead{Funding} European Regional Development Fund (ERDF).

\bmhead{Data Availability} The data analyzed in this study are synthetic and contain no real patient information, provided via GitHub link.

\bmhead{Declaration of generative AI and AI-assisted technologies in the writing process} During the preparation of this work, the authors used ChatGPT and Perplexity AI to assist with language editing and improving the clarity of the manuscript.
}

\bmhead{Acknowledgements}

This article is a result of the project CDMS -- Claim Denial Management Solution, supported by the Innovation and Digital Transition Thematic Programme (COMPETE 2030) and the Lisboa Regional Operational Programme (LISBOA 2030), under the PORTUGAL 2030 Partnership Agreement, through the European Regional Development Fund (ERDF). In addition, part of this work was conducted during the author Sofia Campos' MSc thesis.  

\bmhead{Author Contributions}
All authors contributed to the study conception and design, read, and approved the final manuscript.

\section*{Declarations}

\bmhead{Code Availability}
The code, agent configurations, evaluation scripts, and supporting materials used in this study are publicly available at \url{https://github.com/SofiaCampos03/claims_agentic_hipaa_article}.

\bmhead{Data Availability}
The data analyzed in this study are synthetic and contain no real patient, provider, or payer information. The synthetic dataset and data-generation materials are available through the public repository at \url{https://github.com/SofiaCampos03/claims_agentic_hipaa_article}.

\mycomment{
\bmhead{Institutional Conflicts}
Randy Labs, Effy Business, and the School of Economics and Management, University of Porto.

\bmhead{Individual Conflicts}
Ricardo Macedo (INESC TEC Researcher) and João Tiago Paulo (INESC TEC Researcher).
}

\bmhead{Funding}
This work was supported by the European Regional Development Fund (ERDF) through the CDMS -- Claim Denial Management Solution project.

\bmhead{Declaration of generative AI and AI-assisted technologies in the writing process}
During the preparation of this work, the authors used ChatGPT and Perplexity AI to assist with language editing and improving the clarity of the manuscript.

\clearpage
\appendix
\section{Supplementary Material}
\label{app:supplementary}

\setcounter{table}{0}
\renewcommand{\thetable}{A\arabic{table}}

\begin{table}[htbp]
  \centering
  \caption{XGBoost hyperparameter search space explored with Optuna and the selected configuration.}
  \label{tab:hyperparameters}
  \begin{tabular*}{\textwidth}{@{\extracolsep{\fill}}>{\hspace{0.8em}\normalfont\ttfamily}l l l >{\normalfont}r@{}}
    \toprule
    \multicolumn{1}{@{}l}{\textbf{Hyperparameter}} & \textbf{Type} & \textbf{Search space} & \textbf{Selected} \\
    \midrule
    \multicolumn{4}{@{}l}{\textbf{Ensemble and tree structure}} \\*[4pt]
    n\_estimators      & integer          & $\{200, 300, \ldots, 600\}$ & 400 \\
    max\_depth         & integer          & $\{3, 4, \ldots, 7\}$       & 6 \\
    min\_child\_weight & integer          & $\{1, 2, \ldots, 5\}$       & 2 \\
    \addlinespace[5pt]
    \multicolumn{4}{@{}l}{\textbf{Learning rate and sampling}} \\*[4pt]
    learning\_rate     & continuous (log) & $[0.005, 0.2]$              & 0.0119 \\
    subsample          & continuous       & $[0.6, 1.0]$                & 0.671 \\
    colsample\_bytree  & continuous       & $[0.6, 1.0]$                & 0.906 \\
    \addlinespace[5pt]
    \multicolumn{4}{@{}l}{\textbf{Regularisation}} \\*[4pt]
    gamma              & continuous       & $[0, 0.4]$                  & 0.356 \\
    reg\_alpha         & continuous       & $[0, 2]$                    & 1.765 \\
    reg\_lambda        & continuous       & $[0, 2]$                    & 0.176 \\
    \addlinespace[5pt]
    \multicolumn{4}{@{}l}{\textbf{Class imbalance}} \\*[4pt]
    scale\_pos\_weight & categorical      & $\{1,\ n_{\text{neg}}/n_{\text{pos}} \approx 2.65\}$ & 1 \\
    \bottomrule
  \end{tabular*}
\end{table}

\renewcommand{\thetable}{A\arabic{table}}

{\scriptsize
\renewcommand{\arraystretch}{1.15}
\setlength{\tabcolsep}{4pt}
\begin{xltabular}{\textwidth}{@{}>{\hspace{0.8em}}l l >{\raggedright\arraybackslash\hsize=0.85\hsize}X >{\raggedright\arraybackslash\hsize=1.15\hsize}X@{}}
\caption{Variables used by the framework: type, description and model encoding}
\label{tab:appendix_variables} \\
\toprule
\multicolumn{1}{@{}l}{\textbf{Variable}} & \textbf{Type} & \textbf{Description} & \textbf{Encoding / use} \\
\midrule
\endfirsthead

\caption[]{Variables used by the framework (continued)} \\
\toprule
\multicolumn{1}{@{}l}{\textbf{Variable}} & \textbf{Type} & \textbf{Description} & \textbf{Encoding / use} \\
\midrule
\endhead

\midrule
\multicolumn{4}{r}{\textit{Continued on next page}} \\
\endfoot

\bottomrule
\endlastfoot

\multicolumn{4}{@{}l}{\textbf{Identity and outcome}} \\*[4pt]
\texttt{claim\_id} & integer & Unique claim identifier & Key; pseudonymised in reports \\
\texttt{claim\_status} & categorical & Final status: paid, partially paid, rejected & Target (rejected or partially paid $\rightarrow$ 1) \\
\addlinespace[5pt]

\multicolumn{4}{@{}l}{\textbf{Claim}} \\*[4pt]
\texttt{claim\_total} & continuous & Billed amount (USD) & Box-Cox + z-score; recourse, statistics \\
\texttt{claim\_type} & categorical (5) & Institutional, professional, oral, pharmacy, vision & One-hot \\
\texttt{claim\_subtype} & categorical (2) & Inpatient / outpatient & One-hot \\
\texttt{claim\_month} & categorical & Month of the claim & One-hot \\
\texttt{claim\_day\_of\_week} & categorical & Weekday of the claim & One-hot \\
\texttt{claim\_total\_per\_diagnosis} & continuous & Amount per diagnosis & 5 fixed bins \\
\addlinespace[5pt]

\multicolumn{4}{@{}l}{\textbf{Administrative process}} \\*[4pt]
\texttt{bureaucracy\_time} & interval & Days before adjudication & z-score; recourse lever \\
\texttt{has\_pre\_auth} & binary & Pre-authorization requested & One-hot + \texttt{no\_pre\_auth\_flag}; recourse lever \\
\texttt{was\_pre\_auth\_approved} & binary & Pre-authorization approved & One-hot + \texttt{pre\_auth\_denied\_flag} \\
\addlinespace[5pt]

\multicolumn{4}{@{}l}{\textbf{Payer and patient history}} \\*[4pt]
\texttt{insurer\_name} & categorical (11) & Payer (fictional names) & One-hot; pseudonymised in reports \\
\texttt{insurer\_avg\_claim\_total} & continuous & Payer's average claim amount & Box-Cox + z-score \\
\texttt{patient\_claim\_count} & count & Number of patient claims & z-score \\
\texttt{patient\_avg\_claim\_total} & continuous & Patient's average claim amount & Box-Cox + z-score \\
\texttt{patient\_rejection\_ratio\_120d} & ratio [0,1] & Patient's denial ratio, previous 120 days & 5 fixed bins + flag ($\geq 0.5$) \\
\addlinespace[5pt]

\multicolumn{4}{@{}l}{\textbf{Demographics}} \\*[4pt]
\texttt{age\_range} & ordinal (8) & Age band (0--5 \dots\ 65+) & One-hot \\
\texttt{gender} & categorical (2) & Female / male & One-hot \\
\addlinespace[8pt]

\multicolumn{4}{@{}l}{\textbf{Clinical codes}} \\*[4pt]
\texttt{diagnosis\_codes} & code list & ICD-10 diagnoses & Multi-hot (top 50, support $\geq 10$); code graph / GNN \\
\texttt{procedure\_codes} & code list & Procedures / service items & Multi-hot (top 50, support $\geq 10$); code graph / GNN \\
\texttt{medication\_codes} & code list & Medications (empty for 41.9\% of claims) & Multi-hot (top 50, support $\geq 10$); code graph / GNN \\
\end{xltabular}
}

\clearpage

\begin{center}
\captionof{table}{Exploratory data analysis ($n = 99{,}079$; overall denial rate 27.4\%)}
\label{tab:appendix_eda}
\scriptsize
\renewcommand{\arraystretch}{1.2}
\setlength{\tabcolsep}{4pt}

\begin{tabular*}{\textwidth}{@{\extracolsep{\fill}} p{0.34\textwidth} r r r r @{}}
\toprule
\multicolumn{5}{@{}l}{\textit{(a) Continuous and count variables}} \\
\addlinespace[2pt]
 & & \multicolumn{2}{c}{\textbf{Median}} & \\
\cmidrule(lr){3-4}
\textbf{Variable} & \textbf{Skewness} & \textbf{Denied} & \textbf{Paid} & \textbf{Spearman $\rho$} \\
\midrule
Patient denial ratio, 120 days$^{\dagger}$ & 2.31 & 0 & 0 & 0.56 \\
Bureaucracy time (days) & 1.49 & 43.3 & 18.0 & 0.46 \\
Claim amount (USD) & 1.79 & 3,159 & 1,330 & 0.26 \\
Patient's average amount (USD) & 1.80 & 3,300 & 1,456 & 0.25 \\
Number of medications & 1.11 & 1 & 1 & 0.15 \\
Number of procedures & 0.12 & 5 & 5 & 0.05 \\
Payer's average amount (USD) & 7.88 & 2,985 & 2,983 & 0.02 \\
Number of diagnoses & 0.05 & 3 & 3 & 0.00 \\
Patient claim count & 2.33 & 1 & 1 & 0.00 \\
\midrule
\end{tabular*}

\begin{tabular*}{\textwidth}{@{\extracolsep{\fill}} p{0.34\textwidth} r >{\raggedright\arraybackslash}p{0.45\textwidth} @{}}
\multicolumn{3}{@{}l}{\textit{(b) Categorical variables}} \\
\addlinespace[2pt]
\textbf{Variable} & \textbf{Cram\'er's V} & \textbf{Denial rate by category} \\
\midrule
Claim type & 0.22 & oral~70.8\%\catsep pharmacy~25.8\%\catsep professional~25.4\%\newline institutional~25.1\%\catsep vision~25.1\% \\
Pre-authorization requested & 0.10 & yes~39.4\% (11.1\% of claims)\catsep no~25.9\% \\
Pre-authorization approved & --- & yes~15.4\%\catsep no / not requested~28.4\% \\
Payer (11) & 0.05 & range 5.7\%--28.5\% \\
Age band (8) & 0.01 & range 27.2\%--28.4\% \\
Claim setting & 0.00 & inpatient~27.2\%\catsep outpatient~27.5\% \\
Gender & 0.00 & female~27.4\%\catsep male~27.5\% \\
\midrule
\end{tabular*}

\begin{tabular*}{\textwidth}{@{\extracolsep{\fill}} p{0.34\textwidth} r r r r @{}}
\multicolumn{5}{@{}l}{\textit{(c) Denial rate by bureaucracy-time quartile}} \\
\addlinespace[2pt]
 & \textbf{Q1} & \textbf{Q2} & \textbf{Q3} & \textbf{Q4} \\
\midrule
Range (days) & $\leq$13.3 & 13.3--20.3 & 20.3--32.2 & $>$32.2 \\
Denial rate & 13.3\% & 11.3\% & 12.5\% & \textbf{72.8\%} \\
\bottomrule
\end{tabular*}

\vspace{4pt}
\begin{minipage}{\textwidth}
\footnotesize\raggedright
$^{\dagger}$Mean 0.12; most patients have no prior denials in the 120-day window.
\end{minipage}
\end{center}

\clearpage

\renewcommand{\thetable}{A\arabic{table}}
\begin{table}[htbp]
  \centering
  \scriptsize
  \renewcommand{\arraystretch}{1.15}
  \setlength{\tabcolsep}{4pt}
  \caption{Evaluation of the agentic framework across planning, evidence, retrieval,
  privacy, compliance, readability, prediction, recourse and cost dimensions.}
  \label{tab:appendix_evaluation}
  \begin{tabularx}{\textwidth}{@{}>{\hspace{0.8em}\raggedright\arraybackslash}X r l l@{}}
    \toprule
    \multicolumn{1}{@{}l}{\textbf{Metric}} & \textbf{Value} & \textbf{95\% CI / uncertainty} & \textbf{Denominator / scope} \\
    \midrule
    \multicolumn{4}{@{}l}{\textbf{Planning}} \\*[4pt]
    Coordinator-selected mandatory steps        & 100\%      & 96.3--100\%\textsuperscript{a}  & 100 claims \\
    Claims requiring quality-gate completion    & 0/100      & $\leq 3\%$\textsuperscript{b}   & 100 claims \\
    \addlinespace[5pt]
    \multicolumn{4}{@{}l}{\textbf{Evidence}} \\*[4pt]
    Mean numerical agreement per report         & 98\%       & not computed\textsuperscript{c} & 100 reports \\
    \addlinespace[5pt]
    \multicolumn{4}{@{}l}{\textbf{Retrieval}} \\*[4pt]
    Policy searches per claim                   & 3.3        & ---                             & 100 claims \\
    Valid policy citations                      & 214/216    & 96.7--99.7\%\textsuperscript{a} & 216 citations \\
    \addlinespace[5pt]
    \multicolumn{4}{@{}l}{\textbf{Privacy} (draft $\rightarrow$ released)} \\*[4pt]
    Payer names                                 & 138 $\rightarrow$ 0 & $\leq 3\%$\textsuperscript{b} & 100 reports \\
    Raw claim IDs                               & 141 $\rightarrow$ 0 & $\leq 3\%$\textsuperscript{b} & 100 reports \\
    Scanner-detected PHI patterns               & 1 $\rightarrow$ 0   & $\leq 3\%$\textsuperscript{b} & 100 reports \\
    \addlinespace[5pt]
    \multicolumn{4}{@{}l}{\textbf{Compliance}} \\*[4pt]
    Reviewer-assessed full compliance           & 98/100     & 93.0--99.4\%\textsuperscript{a} & 100 reports \\
    \addlinespace[5pt]
    \multicolumn{4}{@{}l}{\textbf{Readability}} \\*[4pt]
    Reports without flagged technical jargon    & 23/100     & 15.8--32.2\%\textsuperscript{a} & 100 reports \\
    \addlinespace[5pt]
    \multicolumn{4}{@{}l}{\textbf{Prediction}} \\*[4pt]
    Correctly identified actually denied claims & 81/100     & 72.2--87.5\%\textsuperscript{a} & 100 high-risk claims \\
    \addlinespace[5pt]
    \multicolumn{4}{@{}l}{\textbf{Recourse}} \\*[4pt]
    Claims with a predicted decision flip       & 63/100     & 53.2--71.8\%\textsuperscript{a} & 100 high-risk claims \\
    \addlinespace[5pt]
    \multicolumn{4}{@{}l}{\textbf{Cost}} \\*[4pt]
    Mean latency                                & 17\,s      & ---                             & 100 claims \\
    Specialist data-tool calls per claim        & 5.0        & fixed by design\textsuperscript{d} & 100 claims \\
    \bottomrule
  \end{tabularx}

  \smallskip
  {\raggedright
  \textsuperscript{a}\,Wilson score interval.
  \textsuperscript{b}\,One-sided 95\% upper bound for 0 events out of 100 (rule of three; exact value 2.95\%).
  \textsuperscript{c}\,Report-level mean; per-statement counts were not recorded.
  \textsuperscript{d}\,One call each to the prediction, combinations and recourse tools, and two to the
  explainability tools; outputs are cached per claim, so no step repeats.\par}
\end{table}

\clearpage
\bibliography{sn-bibliography}

\end{document}